\pdfoutput=1
\documentclass{article}

\usepackage[main, final]{neurips_2026}

\usepackage[utf8]{inputenc}
\usepackage[T1]{fontenc}
\usepackage{hyperref}
\usepackage{url}
\usepackage{booktabs}
\usepackage{amsfonts}
\usepackage{nicefrac}
\usepackage{microtype}
\usepackage{xcolor}
\usepackage{colortbl}

\usepackage{amsmath}
\usepackage{amssymb}
\usepackage{mathtools}
\usepackage{amsthm}
\usepackage{bbm}
\usepackage{enumitem}
\usepackage{thm-restate}
\usepackage{multirow}

\usepackage[capitalize,noabbrev]{cleveref}
\usepackage{graphicx}

\theoremstyle{plain}
\newtheorem{theorem}{Theorem}
\newtheorem{proposition}[theorem]{Proposition}

\newtheorem{corollary}[theorem]{Corollary}
\theoremstyle{definition}
\newtheorem{definition}[theorem]{Definition}

\theoremstyle{remark}
\newtheorem{remark}{Remark}

\newcommand{\E}{\mathbb{E}}

\newcommand{\Cov}{\mathrm{Cov}}
\newcommand{\norm}[1]{\left\lVert #1 \right\rVert}

\newcommand{\R}{\mathbb{R}}
\newcommand{\Id}{\mathrm{I}}

\graphicspath{{figures/}}

\definecolor{teal}{rgb}{0.040, 0.470, 0.390}
\hypersetup{colorlinks, citecolor=teal, linkcolor=teal}

\title{Analyzing Learning-Dynamics Across Loss and Prediction Choices for Diffusion Models}

\author{%
  Dongwoo Kim \\
  CSE \& GSAI, POSTECH, South Korea
}

\begin{document}

\maketitle

\begin{abstract}
Denoising generative models predict clean data ($x$), noise ($\epsilon$), or velocity ($v$), while training losses can be measured in any of these spaces.
Although the resulting nine combinations are algebraically related, they can yield substantially different denoising accuracy and sample quality.
We disentangle the effects of prediction and loss spaces.
For a fixed prediction target, we show that changing the loss space is exactly equivalent to timestep reweighting and provide a heuristic analysis of how this reweighting affects generation quality.
We then show that a rank-$m$ linear bottleneck compressing a $D$-dimensional noisy input imposes an irreducible squared-error floor of $D-m$ on noise and velocity prediction, whereas clean-data prediction can exploit low-dimensional data structure.
Combining this bound with prediction-dependent error conversion yields distinct necessary conditions on bottleneck capacity.
Experiments with U-Net and JiT across CIFAR-10, Imagenette, and ImageNet-256 support these theoretical predictions.
Under compression, prediction choice strongly affects both denoising accuracy and sample quality, with noise-prediction errors closely matching the theoretical lower bound.
Our results provide a quantitative understanding of how architectural bottlenecks interact with prediction and loss choices in denoising generative models.
\end{abstract}

\section{Introduction}

Diffusion and flow-based generative models train a neural network to reverse a corruption process: given a noisy observation $z_t$ at noise level $t$, the network predicts a target quantity (clean data $x$, noise $\epsilon$, or velocity $v$) from which the original signal can be recovered \citep{sohl2015deep,ho2020ddpm,song2019score,lipman2023flow,liu2023rectified}.
Since $(x, \epsilon, v)$ are related by affine transforms conditioned on $(z_t, t)$, these parameterizations are algebraically equivalent in the oracle regime, and prior work often treats them as interchangeable up to loss reweighting \citep{salimans2022distillation,karras2022edm,kingma2023understanding}.

Yet algebraic equivalence does not guarantee comparable performance in finite-capacity networks.
\citet{li2025denoise} introduce JiT, a pixel-space Transformer with low-rank patch embeddings, and report substantially better FID with $x$-prediction than with $\epsilon$- or $v$-prediction.
They motivate $x$-prediction through the low-dimensional structure of natural images, suggesting that architectural compression may affect prediction targets differently.
However, how such compression affects the relative difficulty of different prediction targets remains insufficiently understood.

We revisit this question by separating two design choices: the \emph{prediction space} (what the network predicts) and the \emph{loss space} (where errors are measured), considering all nine combinations over $\{x,\epsilon,v\}$.
Although prior work has examined prediction parameterizations and loss weightings jointly~\citep{li2025denoise,gagneux2026training}, the effects of architectural information loss and prediction-dependent error conversion remain less understood.
We analyze loss-space choices through timestep reweighting and prediction-space choices through bottleneck-induced error bounds and conversion amplification.
Together, these analyses yield testable predictions about how architectural compression affects denoising accuracy across prediction targets.

\paragraph{Contributions.}
We summarize our contributions as follows.
\begin{enumerate}
    \item \textbf{A joint $3{\times}3$ analysis of prediction and loss spaces.}
    We show that changing the loss space for a fixed prediction target is exactly equivalent to deterministic timestep reweighting (\cref{thm:weighted_reduction}), collecting known affine identities into a unified formulation.
    We further provide a heuristic analysis of how this reweighting affects generation quality, suggesting that $\epsilon$- and $v$-loss can outperform $x$-loss under high-$t$ sensitivity assumptions (\cref{thm:loss_bound}).

    \item \textbf{Irreducible prediction errors under architectural bottlenecks.}
    For a rank-$m$ linear bottleneck compressing a $D$-dimensional noisy input, we establish an error floor of $D-m$ for $\epsilon$- and $v$-prediction, whereas $x$-prediction can exploit the low-dimensional structure of the data (\cref{thm:nullspace_lb,cor:separation}).
    This result quantifies the information-loss mechanism underlying the prediction-space asymmetry observed in JiT~\citep{li2025denoise}.

    \item \textbf{Prediction-dependent error amplification.}
    We characterize how converting different prediction targets to clean-data estimates scales their errors, with factors $(1-t)^2/t^2$ for $\epsilon$-prediction and $(1-t)^2$ for $v$-prediction (\cref{prop:conversion_amp}).
    Combining these identities with the bottleneck error floor yields prediction-dependent necessary conditions on bottleneck capacity (\cref{cor:pred_threshold}).

    \item \textbf{Empirical validation across architectures and scales.}
    We evaluate prediction and loss choices using synthetic data, U-Net, and JiT across CIFAR-10, Imagenette, and ImageNet-256.
    Our experiments support the predicted effects of loss-space reweighting and quantitatively validate the bottleneck error bounds.
    In particular, the $\epsilon$-prediction error closely matches the theoretical floor under strong compression, while bottleneck-width sweeps reveal distinct capacity requirements across prediction targets.
\end{enumerate}

\section{Related work}

\paragraph{Prediction parameterizations and loss weighting.}
Common diffusion parameterizations predict
noise~\citep{ho2020ddpm}, the score~\citep{song2020score},
or a velocity-related quantity.
\citet{salimans2022distillation} study prediction
parameterizations and loss weightings jointly.
EDM~\citep{karras2022edm} introduces preconditioning that normalizes
the scales of network inputs and regression targets across
noise levels.
\citet{kingma2023understanding} express common diffusion
objectives as weighted losses and connect monotonic weightings
to ELBO maximization with noisy data augmentation.
Explicit weighting strategies include
Min-SNR~\citep{hang2024minsnr} and P2~\citep{choi2022p2},
while \citet{chen2023importance} study noise-schedule design.
For our linear interpolation,
\cref{thm:weighted_reduction} collects the established
loss-conversion identities into a common statement
for a fixed prediction space.

\paragraph{Joint analyses of prediction and loss.}
JiT~\citep{li2025denoise}, a pixel-space Transformer using $x$-prediction with $v$-loss, reports substantial performance differences across prediction targets, motivating a systematic analysis of prediction and loss choices.
Their ablation evaluates all nine combinations of prediction and loss spaces.
\citet{kumar2025loss} compare native-target squared-error objectives and their ELBO-derived counterparts across data, noise, velocity, and score parameterizations, and analyze the rescalings relating these objectives.
\citet{gagneux2026training} systematically study the interplay between output parameterization and loss weighting, comparing denoising PSNR across noise levels and FID under different architectures and data regimes.
Their inverse-variance analysis further provides a statistical motivation for emphasizing low-noise timesteps.

\paragraph{Dimensionality, bottlenecks, and error conversion.}
\citet{jin2026revisiting} analyze the learning dynamics of linear diffusion predictors by decomposing their behavior into components parallel and orthogonal to the data subspace.
They characterize how data dimensionality affects the optimal prediction target and propose $k$-Diff, which learns a data-dependent prediction parameter.
We study predictions formed from a rank-$m$ linear compression
of a $D$-dimensional noisy input.
Under the stated assumptions, we derive an expected squared
$\ell_2$ error floor of $D-m$ for noise and velocity prediction
(\cref{thm:nullspace_lb,cor:separation}).
Prior work has discussed numerical instability and error amplification associated with converting noise predictions to clean-data estimates~\citep{salimans2022distillation,karras2022edm,hoogeboom2023simple}.
Under linear interpolation, we combine the corresponding error-scaling identities with our bottleneck bounds to obtain prediction-dependent necessary conditions on bottleneck rank (\cref{prop:conversion_amp,cor:pred_threshold}).

\section{Setting and notation}

We consider training data $x \sim p_{\mathrm{data}}$ in $\R^D$ and noise $\epsilon \sim \mathcal{N}(0,\Id_D)$.
For $t \in (0,1)$, define the noisy observation
\begin{equation}
  z_t = t x + (1-t)\epsilon,
  \label{eq:linear_schedule}
\end{equation}
and the velocity
\begin{equation}
  v = x - \epsilon.
  \label{eq:velocity_def}
\end{equation}
A model is parameterized by a neural network whose \emph{direct output} (prediction space) is one of
\[
u \in \{x,\epsilon,v\}, \qquad u_\theta(z_t,t) = \mathrm{net}_\theta(z_t,t),
\]
where $\mathrm{net}_\theta(z_t,t)$ is modeled as a trainable neural network parameterized by $\theta$.
Given $(z_t,t)$ and one predicted quantity, the other two are obtained by solving the linear system induced by
\eqref{eq:linear_schedule} and \eqref{eq:velocity_def}. This yields deterministic affine transforms among
$\{x_\theta,\epsilon_\theta,v_\theta\}$.

Training uses a squared loss in a chosen \emph{loss space} $s \in \{x,\epsilon,v\}$:
\begin{equation}
  \mathcal{L}_{s\mid u}(\theta)
  \;=\;
  \E\Big[\norm{s_\theta(z_t,t)-s}^2\Big],
  \label{eq:general_loss}
\end{equation}
where $s_\theta$ is computed from $u_\theta$ by the appropriate transform.
Note that $v$-prediction determines the score: by Tweedie's formula, $\nabla_{z_t}\!\log p(z_t) = \big(t\,v^{\star}(z_t,t) - z_t\big)/(1{-}t)$~\citep{song2020score}, an affine map whose time-dependent scaling corresponds to a specific loss weighting within our framework.

\paragraph{Goal.}
In the oracle regime, all $(s,u)$ are equivalent up to deterministic transformations.
The scientific question is to formalize how \emph{learning dynamics} differ for finite capacity and gradient-based optimization,
and to design experiments that disentangle (i) timestep reweighting effects from (ii) geometric and capacity effects.

\section{Learning dynamics of diffusion}

We analyze how prediction and loss choices affect learning through two distinct mechanisms.
We first establish that changing the loss space induces deterministic timestep reweighting and examine how this weighting affects generation quality.
We then turn to the prediction-space axis, studying how architectural bottlenecks introduce irreducible errors and how prediction-dependent conversion amplifies these errors. Proofs for the theoretical results are collected in \cref{sec:proofs}.

\begin{restatable}[Equivalence of loss spaces as timestep reweighting]{proposition}{weightedreduction}
\label{thm:weighted_reduction}
Assume the linear interpolation and velocity definition
\[
z_t = t x + (1-t)\epsilon,\qquad v=x-\epsilon,\qquad t\in(0,1),
\]
with $(x,\epsilon)$ sampled from the data and noise distributions.
Fix a \emph{prediction space} $u\in\{x,\epsilon,v\}$, meaning the network directly outputs
\[
u_\theta(z_t,t) = \mathrm{net}_\theta(z_t,t).
\]
For any \emph{loss space} $s\in\{x,\epsilon,v\}$, define $s_\theta(z_t,t)$ as the quantity obtained by solving the
linear system induced by $(z_t,t)$ together with the constraints defining the pair $(x,\epsilon,v)$,
with $u$ replaced by $u_\theta$.

Then for every pair $(s,u)\in\{x,\epsilon,v\}^2$ there exist:
\begin{itemize}
\item a scalar slope $a_{s\mid u}(t)\neq 0$ that depends \emph{only} on $t$, and
\item an intercept term $b_{s\mid u}(z_t,t)$ that depends on $(z_t,t)$ but is \emph{independent of $u$},
\end{itemize}
such that the \emph{true} target and the \emph{model} prediction satisfy the aligned affine relations
\begin{equation*}
  s \;=\; a_{s\mid u}(t)\,u + b_{s\mid u}(z_t,t),
\end{equation*}
and
\begin{equation*}
  s_\theta(z_t,t) \;=\; a_{s\mid u}(t)\,u_\theta(z_t,t) + b_{s\mid u}(z_t,t).
  \label{eq:aligned_affine}
\end{equation*}
Consequently, the residuals obey the pointwise scaling identity
\begin{equation}
  s_\theta(z_t,t)-s \;=\; a_{s\mid u}(t)\,\big(u_\theta(z_t,t)-u\big),
  \label{eq:residual_scaling_rewrite}
\end{equation}
and hence the squared loss in $s$-space is \emph{exactly} a weighted squared loss in the direct output $u$-space:
\begin{align}
  \mathcal{L}_{s\mid u}(\theta)
  &\;:=\;
  \mathbb{E}\big[\|s_\theta(z_t,t)-s\|^2\big] \notag \\
  &\;=\;
  \mathbb{E}\Big[w_{s\mid u}(t)\,\|u_\theta(z_t,t)-u\|^2\Big],
  \label{eq:loss_weighted_u}
\end{align}
where $w_{s\mid u}(t) := a_{s\mid u}(t)^2 > 0$.
Since $w_{s\mid u}$ does not depend on $\theta$, the Bayes-optimal predictor $u^\star(z_t,t) = \E[u \mid z_t,t]$ is the same for all loss spaces $s$.
For the linear schedule, the weights are listed in \cref{tab:weights}; the individual identities appear in prior work~\citep{salimans2022distillation,kingma2023understanding,karras2022edm}, and we collect them here for reference.
\end{restatable}

\Cref{thm:weighted_reduction} states that changing loss space $s$ only reweights timesteps; the Bayes-optimal predictor $u^\star$ is unchanged. Empirical differences must arise from finite capacity and optimization dynamics, not from a different oracle.
In expectation, applying a weight $w(t)$ in the loss is equivalent to sampling $t$ from a tilted distribution $\propto w(t)\,p(t)$.
In practice, off-diagonal weights can diverge (e.g., $w_{v|x}(t) = (1{-}t)^{-2}$); following~\citep{li2025denoise}, we clamp $(1{-}t) \geq 0.05$.
By contrast, changing prediction space $u$ changes $u^\star$ itself, so is \emph{not} covered by this equivalence.

\begin{table}[t!]
\centering
\caption{The implicit timestep weights $w_{s\mid u}(t)$ induced by different loss-space choices, derived from the linear-schedule relations between $(x,\epsilon,v)$ collected from prior work~\citep{salimans2022distillation,kingma2023understanding,karras2022edm}. Diagonal entries ($u{=}s$) are unity. Off-diagonal entries involving $(1{-}t)^{-1}$ or $t^{-1}$ diverge at endpoints and require clamping in practice.}

\begin{tabular}{c|ccc}
\toprule
{Weight} $w_{s\mid u}(t)$ & {Loss} $s=x$ & {Loss} $s=\epsilon$ & {Loss} $s=v$ \\
\midrule
{Pred} $u=x$
& $1$
& $\left(\frac{t}{1-t}\right)^2$
& $\left(\frac{1}{1-t}\right)^2$ \\[1.5ex]
{Pred} $u=\epsilon$
& $\left(\frac{1-t}{t}\right)^2$
& $1$
& $\left(\frac{1}{t}\right)^2$ \\[1.5ex]
{Pred} $u=v$
& $(1-t)^2$
& $t^2$
& $1$ \\
\bottomrule
\end{tabular}
\label{tab:weights}
\end{table}

The weights connect to the SNR framework~\citep{kingma2023understanding}: for $x$-prediction, $w_{\epsilon|x}(t) = \mathrm{SNR}(t) = t^2/(1{-}t)^2$, so $\epsilon$-loss concentrates optimization on high-SNR timesteps while $x$-loss distributes it uniformly.
This subsumes EDM's weighting~\citep{karras2022edm} and explains why $\epsilon$/$v$-loss emphasize the generation-critical high-$t$ regime.

\subsection{Loss-space ordering: Why some weightings are better}
\label{sec:loss_ordering}

\Cref{thm:weighted_reduction} establishes that changing the loss space $s$ (with fixed prediction space $u$) is equivalent to changing the timestep weighting $w_{s|u}(t)$.
At the population level, all weightings yield the same Bayes-optimal predictor $u^{\star}(z_t,t)$.
However, with finite model capacity, the weighting determines which timesteps receive more optimization pressure.
We now provide a heuristic analysis of why this matters for generation quality, then validate the key prediction empirically.

\paragraph{Setup.}
Consider an ODE-based sampler that integrates from $t{=}0$ (noise) to $t{=}1$ (data) using the learned velocity field $\hat{v}(z_t,t)$.
Let $\delta v(t) := \hat{v}(z_t,t) - v^{\star}(z_t,t)$ denote the velocity error at time $t$, and $\mathrm{MSE}_v(t) := \E\|\delta v(t)\|^2$.
By a first-order perturbation of the ODE $dz/dt = v(z_t,t)$, the final-sample error satisfies
\begin{equation}
\label{eq:sample_error}
\delta z_1 \;\approx\; \int_0^1 \delta v(t) \, dt.
\end{equation}
This linearization is accurate when velocity errors are small relative to the signal.
For the heuristic surrogate below, we restrict the integration to an interior timestep interval $\mathcal{T}=[0.05,0.95]$, neglecting contributions outside this interval.
By \cref{thm:weighted_reduction} with $u{=}v$, a loss in space $s$ weights the squared velocity error by the effective weight $w_{\mathrm{eff}}(t) := w_{s\mid v}(t)$, which is $(1{-}t)^2$, $t^2$, and $1$ for $x$-, $\epsilon$-, and $v$-loss (\cref{tab:weights}).
These effective weights do not depend on the prediction space $u$.

\begin{proposition}[Loss-Space Ordering (heuristic)]
\label{thm:loss_bound}
Consider the following modeling assumptions:
\begin{enumerate}[label=(\roman*)]
\item \textbf{First-order approximation:} The sample error is dominated by~\eqref{eq:sample_error}.
\item \textbf{Cross-time correlations:} By (i), $\E\|\delta z_1\|^2 \approx \int_0^1\!\int_0^1 \E[\delta v(t)^\top \delta v(t')]\,dt\,dt'$. We summarize the cross-time correlations by a sensitivity $\kappa(t) \geq 0$, so that $\E\|\delta z_1\|^2 \approx \int_\mathcal{T} \kappa(t)\,\mathrm{MSE}_v(t)\,dt$, and assume that $\kappa$ does not depend on the training weighting.
\item \textbf{Capacity allocation:} With finite model capacity trained under effective weight $w_{\mathrm{eff}}(t)$ and time distribution $p(t)$, the per-timestep velocity MSE scales as
\begin{equation}
\label{eq:capacity_model}
\mathrm{MSE}_v(t) \;\propto\; \frac{1}{p(t)\, w_{\mathrm{eff}}(t)}.
\end{equation}
\end{enumerate}
Under these assumptions, the expected sample error satisfies
\begin{equation}
\label{eq:gen_bound}
\E\!\left[\|\delta z_1\|^2\right]
\;\propto\;
J_\kappa(w_{\mathrm{eff}}) := \int_\mathcal{T} \frac{\kappa(t)}{p(t)\, w_{\mathrm{eff}}(t)} \, dt.
\end{equation}
If in addition
\begin{enumerate}[label=(\roman*),resume]
\item \textbf{High-$t$ sensitivity:} for natural images, $\kappa(t)$ is concentrated at high $t$,
\end{enumerate}
then
\begin{equation}
\label{eq:loss_ordering}
\text{generation quality}(\epsilon\text{-loss}) \;\approx\; \text{generation quality}(v\text{-loss}) \;\gg\; \text{generation quality}(x\text{-loss}).
\end{equation}
\end{proposition}
\noindent\emph{Derivation.} By (i) and (ii), $\E\|\delta z_1\|^2 \approx \int_0^1 \kappa(t)\,\mathrm{MSE}_v(t)\,dt$, and substituting (iii) gives~\eqref{eq:gen_bound}.
Under (iv), $J_\kappa$ is dominated by high $t$.
At high $t$, the $\epsilon$-loss weight $t^2$ approaches the $v$-loss weight $1$, whereas the $x$-loss weight $(1-t)^2$ approaches zero, thereby placing diminishing emphasis on velocity-space errors.
The $\epsilon$- and $v$-loss integrals are therefore close, and the $x$-loss integral is much larger, which is~\eqref{eq:loss_ordering}.

\paragraph{Discussion.}
Assumption (i) approximates the accumulated sampling error by integrating local velocity errors, neglecting their propagation through the evolving trajectory.
Assumptions (ii) and (iii) are heuristics: real networks share parameters across timesteps and the actual allocation depends on architecture, optimizer, and data.
Assumption (iv) is a property of the data rather than of the model.
We state the result as a proposition to make each premise falsifiable; the ordering is heuristic, not universal.
\Cref{sec:loss_integral} evaluates $J_\kappa$ under the time distributions used in training. With constant $\kappa$, the $x$-loss integral already exceeds the other two by one to two orders of magnitude, so (iv) is needed only to place $\epsilon$-loss next to $v$-loss.
Empirically, the ordering holds for $x$- and $v$-prediction on the U-Net and, more weakly, for $x$-prediction on JiT-B/8 (\cref{sec:loss_analysis}).
It fails on simple 2D distributions, where (iv) does not hold (\cref{sec:exp_details}), and under strong compression on ImageNet-256, where $\epsilon$-loss gives the worst FID for $x$-prediction (\cref{tab:imagenet256}).

\medskip

\begin{remark}[Connection to existing strategies]
\label{rem:edm_connection}
Our framework subsumes existing weighting strategies: EDM~\citep{karras2022edm} weights the $x$-space error by our $\epsilon$-loss weight plus a constant (\cref{sec:edm}), min-SNR~\citep{hang2024minsnr} interpolates between uniform and $\epsilon$-loss, and P2~\citep{choi2022p2} modulates by perceptual priority. All of these upweight high-$t$ relative to $x$-loss, consistent with~\eqref{eq:loss_ordering}.
DDPM~\citep{ho2020ddpm} uses $(\epsilon, \epsilon)$, SD3~\citep{esser2024sd3} uses $(v, v)$, JiT~\citep{li2025denoise} uses $(x, v)$.
\end{remark}

\subsection{Bottleneck lower bounds from information loss: why $x$ is special under manifold structure}

We now turn from loss-space reweighting to the effect of prediction-space choices under architectural bottlenecks.
Clean data $x$ lies on or near a low-dimensional manifold of natural images, whereas the Gaussian noise $\epsilon$ and the velocity $v = x - \epsilon$ fill the ambient $\R^D$. We will show that combining this asymmetry with an architectural information bottleneck yields a sharp consequence: $\epsilon$- and $v$-prediction incur an irreducible error floor, while $x$-prediction remains feasible.

\begin{definition}[Bottleneck model]
\label{def:bottleneck}
We use the following assumptions.
\begin{enumerate}[label=(A\arabic*)]
\item \textbf{Subspace data:} $x = P \hat x$, where $P\in \R^{D\times d}$ satisfies $P^\top P = \Id_d$ with $d \ll D$, and $\hat x \sim p_{\hat x}$ on $\R^d$.
\item \textbf{Isotropic noise:} $\epsilon \sim \mathcal{N}(0,\Id_D)$, independent of $x$.
\item \textbf{Linear bottleneck:} the model observes only $\rho = E z_t \in \R^m$, where $E\in\R^{m\times D}$ and $\mathrm{rank}(E)=m < D$.
\end{enumerate}
\end{definition}

(A1) is the linear idealization of the manifold hypothesis. (A3) matches JiT's patch embedding, which applies two linear convolutions with no nonlinearity in between.

For a Vision Transformer, $D$ corresponds to the flattened patch dimension $d_{\mathrm{patch}} = p^2 C$ for $p \times p$ patches with $C$ channels, and $m$ corresponds to the patch-embedding dimension $d_{\mathrm{bn}}$ to which the architecture compresses each patch before processing. We use $d_{\mathrm{patch}}$ and $d_{\mathrm{bn}}$ throughout when referring to these quantities for concrete architectures, and write $r := d_{\mathrm{bn}} / d_{\mathrm{patch}}$ for the bottleneck ratio, where $r < 1$ indicates patch-level compression and $r \geq 1$ no compression.

\begin{restatable}[Irreducible error from null-space components]{theorem}{irreducible}\label{thm:nullspace_lb}
Assume (A3) and let $\Pi_{\mathrm{null}}$ denote the orthogonal projector onto $\mathrm{null}(E)$.
For any measurable estimator $\hat u(\rho,t)$ of a target $u \in \R^D$,
\begin{align}
  \E\big[\norm{\hat u(\rho,t)-u}^2\big]
  &\;\ge\;
  \E\big[\norm{\E[u\mid \rho,t]-u}^2\big] \label{eq:lb_general}\\
  &\;\ge\;
  \E\big[\norm{\Pi_{\mathrm{null}} u - \E[\Pi_{\mathrm{null}} u \mid \rho,t]}^2\big].
   \notag
\end{align}
If in addition (A2) holds, then $\Pi_{\mathrm{null}}\epsilon$ is independent of $(\rho,t)$, and for $u=\epsilon$
\[
\E\big[\norm{\hat \epsilon(\rho,t)-\epsilon}^2\big]
\;\ge\;
\E\big[\norm{\Pi_{\mathrm{null}}\epsilon}^2\big]
=
(D-m).
\]
\end{restatable}

Intuitively, the observation $\rho$ carries no information about the component of $\epsilon$ in $\mathrm{null}(E)$, and that component has expected squared norm exactly $D{-}m$.
No estimator that sees only $\rho$ can recover it, so the null-space residual $\E\|\Pi_{\mathrm{null}} u - \E[\Pi_{\mathrm{null}} u \mid \rho, t]\|^2$ is an \emph{irreducible error floor} from information loss through $E$.
Data on a low-dimensional subspace can pass through the projection intact, while full-dimensional noise cannot. The next corollary makes this precise.

\begin{corollary}[Bottleneck separation between $x$ and $\epsilon$/$v$]\label{cor:separation}
Assume (A1) to (A3) and
\begin{enumerate}[label=(A\arabic*),start=4]
\item \textbf{Injectivity on the data subspace:} $\mathrm{rank}(EP)=d$, which requires $m \geq d$.
\end{enumerate}
Then the bottleneck preserves all directions in the
$d$-dimensional data subspace, since $\mathrm{rank}(EP)=d$.
In contrast, every estimator of
$u\in\{\epsilon,v\}$ based on $(\rho,t)$ satisfies
\[
    \E\|\hat u-u\|^2 \geq D-m.
\]
\end{corollary}

Thus, the bottleneck can preserve the low-dimensional
data subspace while discarding noise components in
its $(D-m)$-dimensional null space.
This creates an irreducible prediction-error floor
for $\epsilon$- and $v$-prediction that does not
arise from the same null-space mechanism for
$x$-prediction.

\paragraph{When does the bottleneck apply?}
U-Nets bypass this via \emph{skip connections} (effective rank $D$), and EDM's $c_{\mathrm{skip}}$~\citep{karras2022edm} provides a direct input-to-output path; \cref{sec:edm} places EDM in our framework.
Architectures with patch-level compression and \emph{no} skip path (JiT~\citep{li2025denoise}, any ViT with $d_{\mathrm{bn}} < d_{\mathrm{patch}}$) satisfy the assumption directly.
Since $E$ is learned end-to-end, the gradient incentivizes it to preserve the data subspace for $x$-prediction, confirmed by the near-identical MSE across loss types in \cref{tab:jit_results}.

\medskip

\begin{remark}[Nonlinearity after linear compression]
\label{rem:nonlinear}
The lower bound of \cref{thm:nullspace_lb} applies
to any predictor that observes the noisy input only
through a rank-$m$ linear projection
$\rho=Ez_t$, regardless of nonlinear processing
applied after this projection.
In particular, inserting a GELU between JiT's
two bottleneck convolutions preserves the initial
linear compression and therefore does not violate
the assumptions of the theorem.
Our ablation in \cref{tab:nonlinear_bottleneck}
shows similar prediction-space asymmetry with
and without this additional nonlinearity.
This demonstrates robustness to nonlinear processing
after compression, rather than extending the
theoretical guarantee to arbitrary nonlinear
compression maps $f:\R^D\to\R^m$.
\end{remark}

\subsection{Conversion amplification: Why $\epsilon$-prediction requires more capacity}
\label{sec:conversion_amplification}

\Cref{cor:separation} gives the same lower bound, $D-m$,
on the expected squared prediction errors for $\epsilon$
and $v$ under a bottleneck.
We show that conversion to $x$-space yields different
lower bounds and, consequently, different necessary
conditions on bottleneck rank.

\begin{restatable}[Conversion amplification]{proposition}{conversionamp}
\label{prop:conversion_amp}
Let $\hat{u}(z_t, t)$ be a model's prediction in space $u \in \{x, \epsilon, v\}$, and let $\hat{x}_u(z_t, t)$ denote the clean-data estimate obtained by algebraic conversion.
If the model incurs pointwise prediction error $\delta_u := \hat{u} - u$ against the true target, then the induced $x$-space error satisfies:
\begin{align}
\|\hat{x}_x - x\|^2 &= \|\delta_x\|^2, \label{eq:amp_x} \\
\|\hat{x}_v - x\|^2 &= (1-t)^2 \|\delta_v\|^2, \label{eq:amp_v} \\
\|\hat{x}_\epsilon - x\|^2 &= \frac{(1-t)^2}{t^2} \|\delta_\epsilon\|^2. \label{eq:amp_eps}
\end{align}
\end{restatable}

The proof follows by direct algebraic conversion: $\hat{x}_v - x = (1{-}t)\delta_v$ and $\hat{x}_\epsilon - x = \frac{-(1{-}t)}{t}\delta_\epsilon$.

Under the bottleneck assumptions, both $\epsilon$- and
$v$-prediction incur an expected squared error of at least
$D-m$ (\cref{cor:separation}).
Combining this error floor with the conversion factors in
\cref{prop:conversion_amp} gives prediction-dependent lower
bounds on the $x$-space error.
The following corollary states the resulting necessary
conditions on bottleneck rank.

\begin{corollary}[Prediction-type-dependent threshold]
\label{cor:pred_threshold}
Under (A1) -- (A4), for any $\epsilon$- or $v$-predictor based on $(\rho,t)$,
the converted $x$-space MSE satisfies
\begin{align}
\mathrm{MSE}_{v\text{-pred}}        &\geq \E_t[(1{-}t)^2] \cdot (D - m)_+, \\
\mathrm{MSE}_{\epsilon\text{-pred}} &\geq \E_t\!\left[\tfrac{(1-t)^2}{t^2}\right] (D - m)_+.
\end{align}
Under logit-normal sampling with $\mu=-0.5$ and $\sigma=1$,
the expected conversion weights are approximately $0.40$
for $v$-prediction and $20$ for $\epsilon$-prediction.
For the same target $x$-space MSE, the necessary upper bound
on $D-m$ is therefore about $50$ times smaller for
$\epsilon$-prediction than for $v$-prediction.
These lower bounds become trivial at full rank ($m=D$).
\end{corollary}

We empirically examine the predicted difference in bottleneck sensitivity in \cref{sec:loss_analysis}.
The bounds in \cref{cor:pred_threshold} become trivial without compression and therefore do not explain the FID differences in that regime.
Nevertheless, the per-timestep analysis in \cref{sec:imagenet256} reveals error patterns consistent with conversion amplification, although a complete explanation of the FID gap remains open.

\section{Experiments}
\label{sec:loss_analysis}

We systematically evaluate all nine combinations of prediction target $u \in \{x, \epsilon, v\}$ and loss space $s \in \{x, \epsilon, v\}$ across two architectures: a standard U-Net with skip connections (${\sim}$97M params, no information bottleneck) and the JiT-B architecture~\citep{li2025denoise} (${\sim}$130M params, with a configurable bottleneck in the patch embedding). Experiments span CIFAR-10 ($32{\times}32$), Imagenette ($64{\times}64$) and ImageNet-256, with synthetic data validating dimensional scaling.

\paragraph{Setup.}
The U-Net is trained unconditionally with Adam ($\beta_1{=}0.9$, $\beta_2{=}0.999$, lr${}=2{\times}10^{-4}$), logit-normal time sampling ($\mu{=}{-}0.5$, $\sigma{=}1.0$), gradient clipping at norm 1.0, and EMA with decay $0.9999$.
JiT-B is trained class-conditionally with AdamW ($\beta_1{=}0.9$, $\beta_2{=}0.95$), logit-normal time ($\mu{=}{-}0.8$, $\sigma{=}0.8$), label dropout $0.1$ for classifier-free guidance (CFG)~\citep{ho2022classifier}, and EMA with decay $0.9999$.
The CIFAR-10 and Imagenette models are trained for 500 epochs.
For ImageNet-256, we train the compressed $3\times3$ configurations for 100 epochs and the uncompressed controls for 600 epochs, as detailed in \cref{sec:imagenet256}.
JiT uses the full architecture from~\citep{li2025denoise} including RMSNorm, RoPE, QK-norm, SwiGLU, and in-context class tokens.
We report validation MSE in $x$-space (averaged over 3 seeds for CIFAR-10 U-Net, 2 seeds for JiT CIFAR-10) and FID (computed with clean-fid~\citep{parmar2022cleanfid}; U-Net: Heun, 100 steps, unconditional; JiT: Heun, 50 steps, CFG).
JiT's \texttt{BottleneckPatchEmbed} applies two linear convolutions \emph{without} an intermediate nonlinearity, so the bottleneck is a rank-$d_{\mathrm{bn}}$ linear projection: precisely the setting of \cref{thm:nullspace_lb}.
We characterize each configuration by the bottleneck ratio $r := d_{\mathrm{bn}} / d_{\mathrm{patch}}$, where $r < 1$ indicates patch-level compression and $r \geq 1$ no compression.

\paragraph{Loss-space effects without bottleneck: generation quality.}
\Cref{thm:loss_bound} predicts that $\epsilon$-loss and $v$-loss should outperform $x$-loss in generation quality by placing greater optimization weight on high-$t$ timesteps.
We focus on $x$- and $v$-prediction, which achieve substantially better FID than $\epsilon$-prediction under the U-Net setting.
The U-Net results in \cref{tab:unet_results} support the predicted loss ordering.
On CIFAR-10, switching from $x$-loss to $\epsilon$-loss reduces FID from 11.79 to 2.20 for $x$-prediction and from 12.51 to 1.12 for $v$-prediction, with similar improvements on Imagenette.
$v$-loss also consistently outperforms $x$-loss in these settings.

\begin{table}[t]
\caption{U-Net results across datasets. Validation MSE in $x$-space and FID ($\downarrow$).}
\label{tab:unet_results}
\centering
\small
\begin{tabular}{@{}ll ccc c ccc@{}}
\toprule
& & \multicolumn{3}{c}{MSE ($\times 10^{-2}$)} & & \multicolumn{3}{c}{FID} \\
\cmidrule(lr){3-5} \cmidrule(l){7-9}
Dataset & Pred$\backslash$Loss & $x$ & $\epsilon$ & $v$ & & $x$ & $\epsilon$ & $v$ \\
\midrule
\multirow{3}{*}{CIFAR-10}
 & $x$            & 5.395 & 5.360 & 5.361 & & 11.79 &  2.20 & 2.73 \\
 & $\epsilon$  & 5.401 & 5.372 & 5.367 & & 16.45 &  25.49 & 5.88 \\
 & $v$            & 5.392 & \textbf{5.356} & 5.359 & & 12.51 & \textbf{1.12} & 1.68 \\
\midrule
\multirow{3}{*}{Imagenette}
 & $x$            & 4.810 & 4.664 & 4.632 & & 53.40 & 12.76 & 21.05 \\
 & $\epsilon$  & n/a\textsuperscript{\dag} & 4.709 & 4.670 & & 85.77 & 95.41 & 100.98 \\
 & $v$            & 4.763 & 4.649 & \textbf{4.630} & & 43.11 & \textbf{7.06} & 14.32 \\
\bottomrule
\end{tabular}

\vspace{2pt}
{\footnotesize \textsuperscript{\dag}One of two runs diverged (NaN), due to unstable gradients caused by the timestep reweighting.}
\end{table}

\begin{figure*}[t!]
    \centering
    \includegraphics[width=\linewidth]{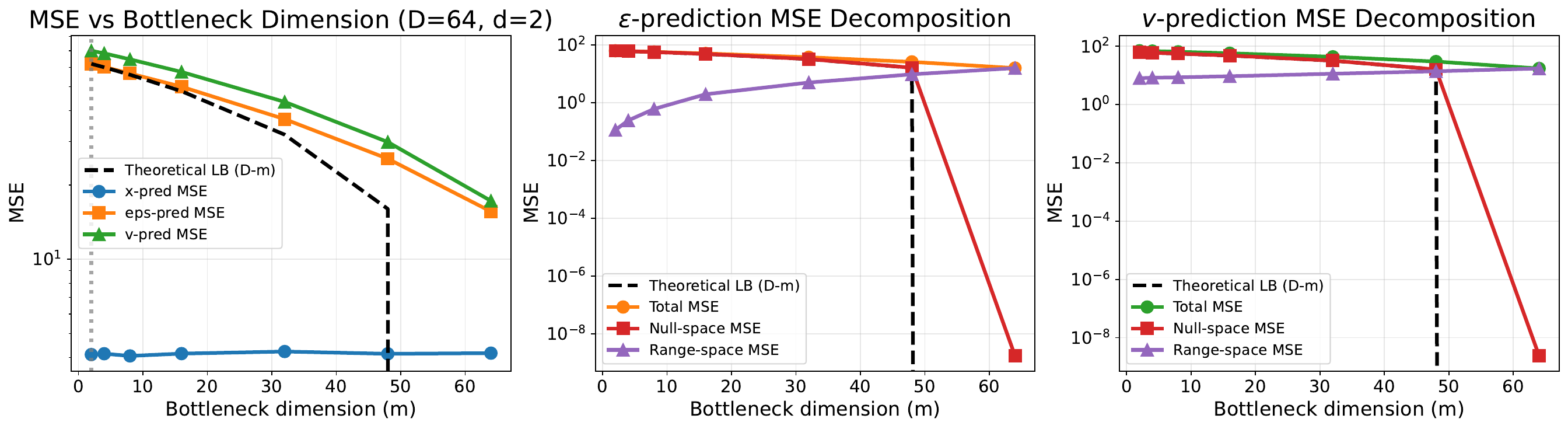}
    \caption{
    Synthetic validation of the bottleneck lower bound
    (\cref{thm:nullspace_lb}) with $D=64$ and $d=2$.
    \textbf{Left:} Prediction MSE in each target's native space
    as a function of bottleneck dimension $m$.
    The $\epsilon$- and $v$-prediction errors are bounded below
    by $D-m$ (dashed), while $x$-prediction maintains low error
    for $m\geq d$.
    \textbf{Middle, Right:} Decomposition of the prediction MSE
    into null-space and range-space components for
    $\epsilon$- and $v$-prediction, respectively.
    }
    \label{fig:bottleneck}
\end{figure*}

\paragraph{Prediction-space effects under bottleneck.}
With an architectural information bottleneck, the prediction target becomes the dominant factor in denoising accuracy.
\Cref{thm:nullspace_lb,cor:separation} establish an irreducible error floor of $D-m$ for $\epsilon$- and $v$-prediction under linear compression.
Our synthetic experiments in \cref{fig:bottleneck} directly validate this bound: the null-space MSE of $\epsilon$-prediction closely matches $D-m$, while that of $v$-prediction remains above the bound.
On Imagenette, \cref{tab:jit_results} shows that JiT-B with bottleneck ratio $r=0.67$ achieves an $x$-space MSE of approximately 0.048 for $x$-prediction, compared with 0.152 for $v$-prediction and 5.85 for $\epsilon$-prediction.
The FID results exhibit the same ordering, while changing the loss space within each prediction type has little effect on MSE.
When the bottleneck is removed ($r=1.33$), the MSE gap narrows substantially: $x$- and $v$-prediction achieve MSE around 0.049, while $\epsilon$-prediction reaches 0.051.
Together, these results show that loss-space reweighting has a relatively modest effect on denoising error, whereas representational information loss can make the prediction target decisive.
The same separation is observed on ImageNet-256 under strong compression ($r=0.17$), where $x$-prediction achieves an $x$-space MSE of 0.023, compared with 0.294 for $v$-prediction and 14.61 for $\epsilon$-prediction (\cref{tab:imagenet256}).
An additional ablation inserting a GELU after
the initial linear compression exhibits a similar
prediction-space asymmetry, indicating robustness
to post-bottleneck nonlinear processing
(\cref{tab:nonlinear_bottleneck}).

\begin{table}[t]
\caption{JiT-B (${\sim}$130M params) on Imagenette 64$\times$64 with class conditioning and CFG. \textbf{Top:} $d_{\mathrm{bn}}{=}128$ ($r=0.67$). \textbf{Bottom:} $d_{\mathrm{bn}}{=}256$ ($r=1.33$). Under compression, $x$-prediction achieves threefold better FID than $v$-prediction and 120-fold better MSE than $\epsilon$-prediction.}
\label{tab:jit_results}
\centering
\small
\begin{tabular}{@{}ll rrr c rrr@{}}
\toprule
& & \multicolumn{3}{c}{MSE ($x$-space)} & & \multicolumn{3}{c}{FID} \\
\cmidrule(lr){3-5} \cmidrule(l){7-9}
Config & Pred$\backslash$Loss & $x$ & $\epsilon$ & $v$ & & $x$ & $\epsilon$ & $v$ \\
\midrule
\multirow{3}{*}{\shortstack{$d_{\mathrm{bn}}{=}128$\\(ratio $0.67$)}}
 & $x$            & 0.050 & \textbf{0.048} & 0.048 & & 41.0 & 37.4 & \textbf{36.0} \\
 & $\epsilon$  & 5.855 & 5.848 & 5.852 & & 146.6 & 147.6 & 151.0 \\
 & $v$            & 0.152 & 0.152 & 0.152 & & 113.2 & 108.3 & 113.0 \\
\midrule
\multirow{3}{*}{\shortstack{$d_{\mathrm{bn}}{=}256$\\(ratio $1.33$)}}
 & $x$            & 0.049 & \textbf{0.048} & 0.048 & & 42.1 & 38.9 & \textbf{37.1} \\
 & $\epsilon$  & 0.051 & 0.051 & 0.051 & & 114.1 & 100.1 & 123.6 \\
 & $v$            & 0.049 & 0.050 & 0.049 & & 58.9 & 60.3 & 58.1 \\
\bottomrule
\end{tabular}
\end{table}

\paragraph{Bottleneck capacity threshold.}
\Cref{cor:pred_threshold} shows that, under compression ($r<1$), the irreducible error floor imposes a stronger constraint on $\epsilon$-prediction than on $v$-prediction, suggesting that $\epsilon$-prediction requires greater bottleneck capacity to achieve comparable denoising accuracy.
Since the bound becomes trivial at $r\geq1$, we empirically investigate recovery by varying the bottleneck width of BottleneckViT, a lightweight ViT that follows JiT's patch-embedding design but adds a GELU after the initial linear projection (see \cref{sec:exp_details}).
On CIFAR-10, the gap between $v$- and $x$-prediction nearly disappears once compression is removed (\cref{tab:bn_sweep_app}).
In contrast, on Imagenette, $\epsilon$-prediction remains less accurate even without compression and approaches $x$-prediction only with a substantially wider embedding ($r\approx4$; \cref{tab:eps_threshold_app}).
This recovery threshold is empirical and is not predicted by \cref{cor:pred_threshold}.
Additional synthetic sweeps over $D$, $d$, and $m$ (\cref{sec:synthetic}) and a patch-size sweep at fixed embedding width (\cref{sec:bottle_sweep}) further support the predicted dependence of prediction errors on $D-m$.

\begin{table}[t!]
\caption{JiT-B/16 on class-conditional ImageNet $256{\times}256$. Validation $x$-MSE and FID ($\downarrow$, 50K samples). \textbf{Top:} $d_{\mathrm{bn}}{=}128$ ($r{=}0.17$). \textbf{Bottom:} $d_{\mathrm{bn}}{=}768$ ($r{=}1.0$). Under compression, prediction type dominates both metrics
($\epsilon$-space MSE $0.834 \approx (D-m)/D = 0.833$;
see \cref{sec:imagenet256}). Without compression and at full budget, MSE converges across prediction types ($0.022$--$0.043$) but FID does not, so aggregate denoising MSE does not determine FID.}
\label{tab:imagenet256}
\centering
\small
\begin{tabular}{@{}ll rrr c rrr@{}}
\toprule
& & \multicolumn{3}{c}{$x$-MSE} & & \multicolumn{3}{c}{FID} \\
\cmidrule(lr){3-5} \cmidrule(l){7-9}
Config & Pred$\backslash$Loss & $x$ & $\epsilon$ & $v$ & & $x$ & $\epsilon$ & $v$ \\
\midrule
\multirow{3}{*}{\shortstack[l]{$d_{\mathrm{bn}}{=}128$\\($r{=}0.17$)}}
 & $x$            & 0.023 & 0.023 & \textbf{0.023} & & 23.9 & 36.8 & \textbf{20.1} \\
 & $\epsilon$  & 14.61 & 14.61 & 14.61 & & 421.2 & 419.0 & 420.1 \\
 & $v$            & 0.294 & 0.294 & 0.294 & & 201.5 & 202.3 & 199.0 \\
\midrule
\multirow{3}{*}{\shortstack[l]{$d_{\mathrm{bn}}{=}768$\\($r{=}1.0$)}}
 & $x$            & --- & --- & \textbf{0.022} & & --- & --- & \textbf{5.7} \\
 & $\epsilon$  & --- & --- & 0.043 & & --- & --- & 330.6 \\
 & $v$            & --- & --- & 0.026 & & --- & --- & 140.6 \\
\bottomrule
\end{tabular}
\end{table}

\paragraph{Convergence behavior without compression.}
The bottleneck lower bound in \cref{thm:nullspace_lb} applies to rank-deficient embeddings ($r<1$) and becomes trivial when $r\geq1$.
On JiT-B/4 CIFAR-10 ($r=2.67$), the initially large denoising MSE gap between $\epsilon$- and $x$-prediction closes with longer training (\cref{fig:convergence_speed}), suggesting that prediction-type differences without compression can partly reflect optimization difficulty rather than irreversible information loss.
However, comparable denoising accuracy does not necessarily imply comparable generation quality.
On ImageNet-256 without compression ($r=1.0$), the models trained for 600 epochs achieve relatively similar $x$-space MSE, yet their FID scores differ substantially (\cref{tab:imagenet256}).

\begin{figure}[t!]
\centering
\includegraphics[width=0.95\linewidth]{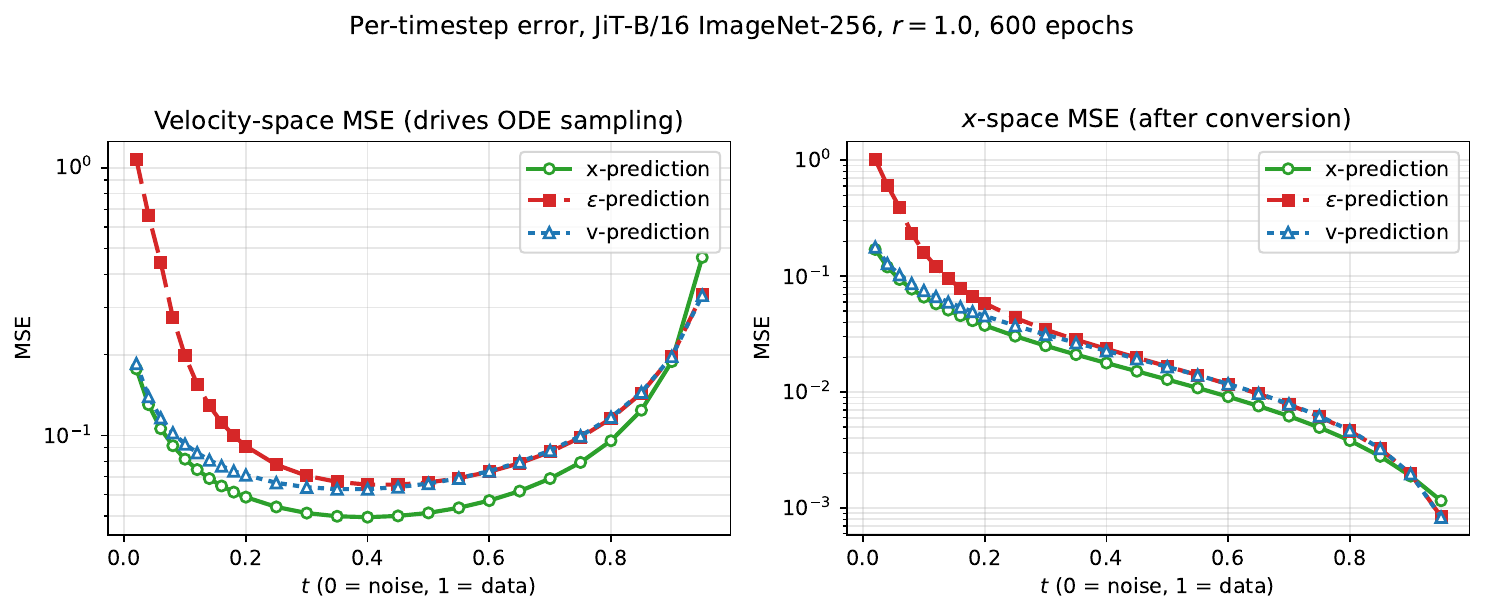}
\caption{
Per-timestep prediction errors on ImageNet-256 using
JiT-B/16 without compression ($r=1.0$, 600 epochs).
\textbf{Left:} Velocity-space MSE against sampled targets, used to assess errors relevant to ODE sampling.
\textbf{Right:} $x$-space MSE after conversion.
The velocity-space error of $\epsilon$-prediction increases
sharply at low $t$, consistent with the $1/t^2$
amplification factor, while $x$- and $v$-prediction
remain bounded over the evaluated timesteps.
Results below $t=0.05$ also reflect denominator clamping.
}
\label{fig:pertimestep}
\end{figure}

To examine the role of error conversion,
\cref{fig:pertimestep} compares per-timestep errors
in velocity space, which is used by ODE sampling,
and in $x$-space after conversion.
The $\epsilon$-prediction model exhibits sharply
increasing velocity-space MSE at low $t$, consistent
with the $1/t^2$ amplification of squared errors
during $\epsilon$-to-velocity conversion.
In contrast, $x$- and $v$-prediction maintain
substantially lower errors in this region.
These observations suggest that conversion
amplification contributes to the observed FID gap,
although it does not fully explain the differences,
particularly the poor FID of $v$-prediction.
\Cref{sec:imagenet256} provides further analyses
in a common velocity space and compares ODE
and SDE sampling.

\section{Conclusion}

We have presented a joint $3{\times}3$ analysis of prediction and loss spaces in denoising generative models.
For a fixed prediction target, changing the loss space is equivalent to timestep reweighting, which can affect generation quality under finite model capacity (\cref{thm:weighted_reduction,thm:loss_bound}).
For prediction-space choices, we show that a rank-deficient linear bottleneck introduces an irreducible error floor of $D-m$ for $\epsilon$- and $v$-prediction, whereas $x$-prediction can exploit low-dimensional data structure (\cref{thm:nullspace_lb,cor:separation}).
Combining this bound with prediction-dependent error conversion yields necessary conditions on bottleneck capacity that are more restrictive for $\epsilon$-prediction than for $v$-prediction (\cref{prop:conversion_amp,cor:pred_threshold}).

Experiments with U-Net and JiT across CIFAR-10, Imagenette, and ImageNet-256 support these analyses.
Under compression, $x$-prediction maintains low denoising error, while $\epsilon$- and $v$-prediction are more sensitive to bottleneck capacity.
In particular, the observed $\epsilon$-prediction error closely matches the theoretical lower bound under strong compression, and bottleneck-width sweeps reveal distinct recovery patterns across prediction targets.
Together, our findings clarify how prediction and loss choices interact with architectural information constraints and provide guidance for selecting prediction targets in compressed generative models.

\paragraph{Limitations.}
Our theoretical bounds rely on a linear bottleneck and a low-dimensional subspace model of the data.
Our GeLU ablation retains the initial linear
bottleneck and therefore does not establish
corresponding guarantees for architectures with general nonlinear compression maps.
The loss-ordering analysis is heuristic and depends on assumptions about sampling sensitivity and the allocation of model capacity across timesteps.

Our experiments span different architectures, conditioning schemes, and datasets, so absolute FID values are not directly comparable across settings.
Moreover, the compressed ImageNet-256 configurations were trained for a shorter budget than the uncompressed controls, and we did not evaluate their convergence under extended training.
Without compression, our error bounds become trivial.
Although prediction-dependent conversion can amplify errors during sampling, particularly for $\epsilon$-prediction, it does not fully explain the observed FID differences.
Our stochastic sampling ablation is also limited to samplers that share the same velocity conversion.
Extending the analysis to latent diffusion, text conditioning, and architectures with explicit skip connections or EDM-style preconditioning remains an important direction for future work.

\clearpage

\begin{ack}
  This work was supported by the National Research Foundation of Korea (NRF) grant funded by the Korea government (MSIT) (RS-2024-00337955); by the Institute of Information \& Communications Technology Planning \& Evaluation (IITP) grants funded by the Korea government (MSIT) (RS-2024-00457882, Artificial Intelligence Research Hub Project; RS-2019-II191906, Artificial Intelligence Graduate School Program (POSTECH)); and by IITP grants funded by the Korea government (MSIT) under the Leading Generative AI Human Resources Development Program (IITP-2026-RS-2026-25546560) and the AI Star Fellowship Support Program (IITP-2026-RS-2026-25616313).
\end{ack}

\bibliographystyle{plainnat}
\bibliography{references}

\begin{thebibliography}{20}
\providecommand{\natexlab}[1]{#1}
\providecommand{\url}[1]{\texttt{#1}}
\expandafter\ifx\csname urlstyle\endcsname\relax
  \providecommand{\doi}[1]{doi: #1}\else
  \providecommand{\doi}{doi: \begingroup \urlstyle{rm}\Url}\fi

\bibitem[Chen(2023)]{chen2023importance}
Ting Chen.
\newblock On the importance of noise scheduling for diffusion models.
\newblock \emph{arXiv preprint arXiv:2301.10972}, 2023.

\bibitem[Choi et~al.(2022)Choi, Lee, Shin, Kim, Kim, and Yoon]{choi2022p2}
Jooyoung Choi, Jungbeom Lee, Chaehun Shin, Sungwon Kim, Hyunwoo Kim, and
  Sungroh Yoon.
\newblock Perception prioritized training of diffusion models.
\newblock In \emph{Proceedings of the IEEE/CVF Conference on Computer Vision
  and Pattern Recognition}, pages 11472--11481, 2022.

\bibitem[Esser et~al.(2024)Esser, Kulal, Blattmann, Entezari, M{\"u}ller,
  Saini, Levi, Lorenz, Sauer, Boesel, Podell, Dockhorn, English, and
  Rombach]{esser2024sd3}
Patrick Esser, Sumith Kulal, Andreas Blattmann, Rahim Entezari, Jonas
  M{\"u}ller, Harry Saini, Yam Levi, Dominik Lorenz, Axel Sauer, Frederic
  Boesel, Dustin Podell, Tim Dockhorn, Zion English, and Robin Rombach.
\newblock Scaling rectified flow transformers for high-resolution image
  synthesis.
\newblock In \emph{International Conference on Machine Learning}, volume 235 of
  \emph{Proceedings of Machine Learning Research}, pages 12606--12633, 2024.

\bibitem[Gagneux et~al.(2026)Gagneux, Martin, Gribonval, and
  Massias]{gagneux2026training}
Anne Gagneux, S{\'e}gol{\`e}ne Martin, R{\'e}mi Gribonval, and Mathurin
  Massias.
\newblock Training flow matching: The role of weighting and parameterization.
\newblock In \emph{ICLR 2026 Workshop on Deep Generative Model in Machine
  Learning: Theory, Principle and Efficacy}, 2026.
\newblock arXiv:2603.06454.

\bibitem[Hang et~al.(2023)Hang, Gu, Li, Bao, Chen, Hu, Geng, and
  Guo]{hang2024minsnr}
Tiankai Hang, Shuyang Gu, Chen Li, Jianmin Bao, Dong Chen, Han Hu, Xin Geng,
  and Baining Guo.
\newblock Efficient diffusion training via min-{SNR} weighting strategy.
\newblock In \emph{Proceedings of the IEEE/CVF International Conference on
  Computer Vision}, pages 7441--7451, 2023.

\bibitem[Ho and Salimans(2022)]{ho2022classifier}
Jonathan Ho and Tim Salimans.
\newblock Classifier-free diffusion guidance.
\newblock \emph{arXiv preprint arXiv:2207.12598}, 2022.

\bibitem[Ho et~al.(2020)Ho, Jain, and Abbeel]{ho2020ddpm}
Jonathan Ho, Ajay Jain, and Pieter Abbeel.
\newblock Denoising diffusion probabilistic models.
\newblock \emph{Advances in neural information processing systems},
  33:\penalty0 6840--6851, 2020.

\bibitem[Hoogeboom et~al.(2023)Hoogeboom, Heek, and
  Salimans]{hoogeboom2023simple}
Emiel Hoogeboom, Jonathan Heek, and Tim Salimans.
\newblock Simple diffusion: End-to-end diffusion for high resolution images.
\newblock In \emph{International Conference on Machine Learning}, pages
  13213--13232, 2023.

\bibitem[Jin and Wang(2026)]{jin2026revisiting}
Qing Jin and Chaoyang Wang.
\newblock Revisiting diffusion model predictions through dimensionality.
\newblock In \emph{Advances in Neural Information Processing Systems}, 2026.

\bibitem[Karras et~al.(2022)Karras, Aittala, Aila, and Laine]{karras2022edm}
Tero Karras, Miika Aittala, Timo Aila, and Samuli Laine.
\newblock Elucidating the design space of diffusion-based generative models.
\newblock \emph{Advances in neural information processing systems},
  35:\penalty0 26565--26577, 2022.

\bibitem[Kingma and Gao(2023)]{kingma2023understanding}
Diederik Kingma and Ruiqi Gao.
\newblock Understanding diffusion objectives as the {ELBO} with simple data
  augmentation.
\newblock \emph{Advances in Neural Information Processing Systems},
  36:\penalty0 65484--65516, 2023.

\bibitem[Kumar et~al.(2025)Kumar, V{\"a}th, and Gregorov{\'a}]{kumar2025loss}
Dibyanshu Kumar, Philipp V{\"a}th, and Magda Gregorov{\'a}.
\newblock Loss functions in diffusion models: A comparative study.
\newblock In \emph{Machine Learning and Knowledge Discovery in Databases.
  Research Track (ECML PKDD 2025)}, Lecture Notes in Computer Science, pages
  190--205. Springer, 2025.
\newblock \doi{10.1007/978-3-032-05981-9_12}.

\bibitem[Li and He(2026)]{li2025denoise}
Tianhong Li and Kaiming He.
\newblock Back to basics: Let denoising generative models denoise.
\newblock In \emph{Proceedings of the IEEE/CVF Conference on Computer Vision
  and Pattern Recognition}, pages 36115--36125, 2026.

\bibitem[Lipman et~al.(2023)Lipman, Chen, Ben-Hamu, Nickel, and
  Le]{lipman2023flow}
Yaron Lipman, Ricky T.~Q. Chen, Heli Ben-Hamu, Maximilian Nickel, and Matt Le.
\newblock Flow matching for generative modeling.
\newblock In \emph{International Conference on Learning Representations}, 2023.

\bibitem[Liu et~al.(2023)Liu, Gong, and Liu]{liu2023rectified}
Xingchao Liu, Chengyue Gong, and Qiang Liu.
\newblock Flow straight and fast: Learning to generate and transfer data with
  rectified flow.
\newblock In \emph{International Conference on Learning Representations}, 2023.

\bibitem[Parmar et~al.(2022)Parmar, Zhang, and Zhu]{parmar2022cleanfid}
Gaurav Parmar, Richard Zhang, and Jun-Yan Zhu.
\newblock On aliased resizing and surprising subtleties in {GAN} evaluation.
\newblock In \emph{Proceedings of the IEEE/CVF Conference on Computer Vision
  and Pattern Recognition}, pages 11410--11420, 2022.

\bibitem[Salimans and Ho(2022)]{salimans2022distillation}
Tim Salimans and Jonathan Ho.
\newblock Progressive distillation for fast sampling of diffusion models.
\newblock In \emph{International Conference on Learning Representations}, 2022.

\bibitem[Sohl-Dickstein et~al.(2015)Sohl-Dickstein, Weiss, Maheswaranathan, and
  Ganguli]{sohl2015deep}
Jascha Sohl-Dickstein, Eric Weiss, Niru Maheswaranathan, and Surya Ganguli.
\newblock Deep unsupervised learning using nonequilibrium thermodynamics.
\newblock In \emph{International conference on machine learning}, pages
  2256--2265. pmlr, 2015.

\bibitem[Song and Ermon(2019)]{song2019score}
Yang Song and Stefano Ermon.
\newblock Generative modeling by estimating gradients of the data distribution.
\newblock \emph{Advances in neural information processing systems}, 32, 2019.

\bibitem[Song et~al.(2021)Song, Sohl-Dickstein, Kingma, Kumar, Ermon, and
  Poole]{song2020score}
Yang Song, Jascha Sohl-Dickstein, Diederik~P Kingma, Abhishek Kumar, Stefano
  Ermon, and Ben Poole.
\newblock Score-based generative modeling through stochastic differential
  equations.
\newblock In \emph{International Conference on Learning Representations}, 2021.

\end{thebibliography}

\newpage
\appendix
\crefalias{section}{appendix}
\crefname{appendix}{Appendix}{Appendices}
\Crefname{appendix}{Appendix}{Appendices}

\section{Proofs}
\label{sec:proofs}
\subsection{Proof of \cref{thm:weighted_reduction}}

\weightedreduction*

\begin{proof}
The result follows from the affine conversion formulas among $(x, \epsilon, v)$ given $(z_t, t)$.
From $z_t = tx + (1{-}t)\epsilon$ and $v = x - \epsilon$, any two of $(x, \epsilon, v)$ determine the third:
\begin{align*}
x &= \frac{z_t - (1{-}t)\epsilon}{t} = z_t + (1{-}t)v, \\
\epsilon &= \frac{z_t - tx}{1{-}t} = z_t - tv, \\
v &= x - \epsilon = \frac{x - z_t}{1{-}t} = \frac{z_t - \epsilon}{t}.
\end{align*}
For any prediction space $u$ and loss space $s$, substituting $u_\theta$ for $u$ in these formulas yields $s_\theta$, and the \emph{same} substitution of the true $u$ yields the true $s$.
Explicitly, the slopes $a_{s|u}(t)$ and intercepts $b_{s|u}(z_t,t)$ are:

\medskip
\noindent

\textbf{Pred $u{=}x$:}
From the linear interpolation, we have
\[
\epsilon
= -\frac{t}{1-t}x + \frac{z_t}{1-t},
\qquad
v
= \frac{x-z_t}{1-t}.
\]
Therefore,
\begin{align*}
a_{\epsilon|x}(t) &= -\frac{t}{1-t},
& b_{\epsilon|x}(z_t,t) &= \frac{z_t}{1-t},\\
a_{v|x}(t) &= \frac{1}{1-t},
& b_{v|x}(z_t,t) &= -\frac{z_t}{1-t}.
\end{align*}

\noindent
\textbf{Pred $u{=}\epsilon$:} \quad $x = (z_t - (1{-}t)\epsilon)/t$, so $a_{x|\epsilon} = -(1{-}t)/t$.
$v = x - \epsilon = (z_t - \epsilon)/t$, so $a_{v|\epsilon} = -1/t$.

\noindent
\textbf{Pred $u{=}v$:} \quad $x = z_t + (1{-}t)v$, so $a_{x|v} = 1{-}t$.
$\epsilon = z_t - tv$, so $a_{\epsilon|v} = -t$.

\medskip
In all cases, $a_{s|u}(t)$ depends only on $t$, and the intercept $b_{s|u}(z_t,t)$ is independent of $u_\theta$ (it involves only $z_t$ and $t$).
Since $s_\theta - s = a_{s|u}(t)(u_\theta - u)$ by linearity, squaring and taking expectations gives
$\mathcal{L}_{s|u} = \E[a_{s|u}(t)^2 \|u_\theta - u\|^2] = \E[w_{s|u}(t) \|u_\theta - u\|^2]$,
where $w_{s|u}(t) = a_{s|u}(t)^2$.
The diagonal entries are $w_{u|u} = 1$, and the off-diagonal entries are:
\begin{align*}
w_{\epsilon|x} &= \left(\tfrac{t}{1{-}t}\right)^{\!2}, \quad w_{v|x} = \left(\tfrac{1}{1{-}t}\right)^{\!2}, \quad w_{x|\epsilon} = \left(\tfrac{1{-}t}{t}\right)^{\!2}, \\
w_{v|\epsilon} &= \left(\tfrac{1}{t}\right)^{\!2}, \quad w_{x|v} = (1{-}t)^2, \quad w_{\epsilon|v} = t^2,
\end{align*}
which are exactly the entries in \cref{tab:weights}.
Since $w_{s|u}(t)>0$ depends only on $t$, the population minimizer over all measurable predictors is
$u^\star(z_t,t)=\E[u\mid z_t,t]$ for every loss space $s$.
\end{proof}

\subsection{Proof of \cref{thm:nullspace_lb}}

\irreducible*

\begin{proof}
Let $\hat{u}(\rho,t)$ be \emph{any} estimator that depends only on the inputs $(\rho,t)$. We define the ideal Bayes estimator $u^\star = \E[u \mid \rho,t]$.
We can decompose the error of $\hat{u}$ as:
\[
    \hat{u} - u \;=\; (\hat{u} - u^\star) + (u^\star - u).
\]
Squaring the norm and taking expectations:
\begin{align*}
    \E\|\hat{u} - u\|^2 &= \E\|\hat{u} - u^\star\|^2 + \E\|u^\star - u\|^2 + 2\E\big[ (\hat{u} - u^\star)^\top (u^\star - u) \big].
\end{align*}
The cross-term vanishes by the tower property of conditional expectation: since $(\hat{u} - u^\star)$ is measurable with respect to $(\rho,t)$, it can be pulled out of the inner expectation:
\[
    \E\big[ (\hat{u} - u^\star)^\top (u^\star - u) \big] = \E_{\rho,t}\Big[ (\hat{u} - u^\star)^\top \underbrace{\E[u^\star - u \mid \rho,t]}_{=0} \Big] = 0.
\]
Thus, we obtain the standard decomposition:
\[
    \E\|\hat{u} - u\|^2 \;=\; \underbrace{\E\|\hat{u} - u^\star\|^2}_{\ge 0} + \E\|u^\star - u\|^2 \;\ge\; \E\|u^\star - u\|^2.
\]
This proves the first inequality.

The term $\E\|u^\star - u\|^2$ represents the total variance of the unrecoverable residual.
For any vector $z$, and any orthogonal projector $\Pi$, we have $\|z\|^2 \ge \|\Pi z\|^2$.
Applying this to the residual vector $z = u - u^\star$ with the null-space projector $\Pi_{\mathrm{null}}$:
\[
    \|u - u^\star\|^2 \;\ge\; \|\Pi_{\mathrm{null}}(u - u^\star)\|^2.
\]
Since $\Pi_{\mathrm{null}}$ is a deterministic linear operator, it commutes with the conditional expectation by linearity of expectation:
\[
    \Pi_{\mathrm{null}}(u - u^\star) = \Pi_{\mathrm{null}}u - \Pi_{\mathrm{null}}\E[u \mid \rho,t] = \Pi_{\mathrm{null}}u - \E[\Pi_{\mathrm{null}}u \mid \rho,t].
\]
Taking expectations proves the second inequality:
\[
    \E\|u^\star - u\|^2 \;\ge\; \E\|\Pi_{\mathrm{null}}u - \E[\Pi_{\mathrm{null}}u \mid \rho,t]\|^2.
\]

Consider the case $u = \epsilon$ with isotropic noise $\epsilon \sim \mathcal{N}(0, \Id)$.
Write $\epsilon = \Pi_{\mathrm{col}}\epsilon + \Pi_{\mathrm{null}}\epsilon$, where $\Pi_{\mathrm{col}} = E^\dagger E$ projects onto the column space of $E$.
Since $\epsilon$ is isotropic Gaussian and $\Pi_{\mathrm{col}}, \Pi_{\mathrm{null}}$ are orthogonal projectors onto complementary subspaces, the components $\Pi_{\mathrm{col}}\epsilon$ and $\Pi_{\mathrm{null}}\epsilon$ are independent Gaussian vectors (a standard property of orthogonal projections of Gaussian vectors).
Since $\rho = Ez_t = E(tx + (1{-}t)\epsilon)$ depends on $\epsilon$ only through $E\epsilon = E\Pi_{\mathrm{col}}\epsilon$ (as $E\Pi_{\mathrm{null}} = 0$ by definition), we have $\Pi_{\mathrm{null}}\epsilon \perp\!\!\!\perp (\rho,t)$.
Therefore:
\[
    \E[\Pi_{\mathrm{null}}\epsilon \mid \rho,t] = \E[\Pi_{\mathrm{null}}\epsilon] = 0.
\]
Substituting this into our bound:
\[
    \E\|\hat{\epsilon} - \epsilon\|^2 \;\ge\; \E\|\Pi_{\mathrm{null}}\epsilon - 0\|^2 \;=\; \E\|\Pi_{\mathrm{null}}\epsilon\|^2.
\]
For isotropic Gaussian noise, the expected squared norm of its projection is simply the dimension of the subspace:
\[
    \E\|\Pi_{\mathrm{null}}\epsilon\|^2 = \mathrm{tr}(\Cov(\Pi_{\mathrm{null}}\epsilon)) = \mathrm{tr}(\Pi_{\mathrm{null}}\Id \Pi_{\mathrm{null}}^\top) = \mathrm{rank}(\Pi_{\mathrm{null}}).
\]
Since $E \in \R^{m \times D}$ has rank $m$, the dimension of its null space is $D-m$. Thus, the error is at least $D-m$.
\end{proof}

\subsection{Proof of \cref{cor:separation}}

\begin{proof}
Under (A1), $x=P\hat{x}$ with $\hat{x}\in\R^d$.
By (A4), $EP\in\R^{m\times d}$ has full column rank, so the bottleneck mapping is injective on the data subspace.
Thus, unlike the noise components in the null space of $E$, the data-subspace components are not irreversibly lost through the linear projection.

For $\epsilon$-prediction, \cref{thm:nullspace_lb} directly gives
\[
    \E\|\hat{\epsilon}-\epsilon\|^2 \geq D-m.
\]

For $v=x-\epsilon$, we have
$\Pi_{\mathrm{null}}v
=\Pi_{\mathrm{null}}x-\Pi_{\mathrm{null}}\epsilon$.
By (A2), $\Pi_{\mathrm{null}}\epsilon$ is independent of $(x,\rho,t)$ and has zero mean.
Therefore,
\[
    \E[\Pi_{\mathrm{null}}v\mid\rho,t]
    =\E[\Pi_{\mathrm{null}}x\mid\rho,t].
\]
Applying \cref{thm:nullspace_lb} and using this independence, we obtain
\begin{align*}
    \E\|\hat{v}-v\|^2
    &\geq
    \E\left\|
        \Pi_{\mathrm{null}}v
        -\E[\Pi_{\mathrm{null}}v\mid\rho,t]
    \right\|^2 \\
    &=
    \E\left\|
        \Pi_{\mathrm{null}}x
        -\E[\Pi_{\mathrm{null}}x\mid\rho,t]
    \right\|^2
    +\E\|\Pi_{\mathrm{null}}\epsilon\|^2 \\
    &\geq D-m.
\end{align*}
\end{proof}

\subsection{Proof of \cref{prop:conversion_amp}}

The conversions from prediction space to $x$-space are: $\hat{x}_x = \hat{x}$, so $\hat{x}_x - x = \delta_x$; $\hat{x}_v = z_t + (1{-}t)\hat{v}$, so $\hat{x}_v - x = (1{-}t)\delta_v$; $\hat{x}_\epsilon = (z_t - (1{-}t)\hat{\epsilon})/t$, so $\hat{x}_\epsilon - x = -(1{-}t)\delta_\epsilon/t$. Taking squared norms yields \eqref{eq:amp_x} through \eqref{eq:amp_eps}.

\subsection{Proof of \cref{cor:pred_threshold}}

Fix $t\in(0,1)$. Under (A2) and (A3),
$\Pi_{\mathrm{null}}\epsilon$ is independent of
$(\rho,t)$. Hence, the arguments of
\cref{thm:nullspace_lb,cor:separation} apply
conditionally on $t$, giving
\[
\E[\|\hat\epsilon-\epsilon\|^2\mid t]
\geq D-m,
\qquad
\E[\|\hat v-v\|^2\mid t]
\geq D-m.
\]
By \cref{prop:conversion_amp}, converting these
prediction errors to $x$-space multiplies them
by $(1-t)^2/t^2$ and $(1-t)^2$, respectively.
Taking expectations over $t$ yields the stated
lower bounds.

For the numerical values, let
$Z=\operatorname{logit}(t)\sim\mathcal N(\mu,\sigma^2)$
with $\mu=-0.5$ and $\sigma=1$.
Since $(1-t)/t=e^{-Z}$,
\[
\E_t\left[\frac{(1-t)^2}{t^2}\right]
=\E[e^{-2Z}]
=e^{-2\mu+2\sigma^2}
=e^3\approx20.1,
\]
while numerical quadrature gives
\[
\E_t[(1-t)^2]\approx0.403.
\]
The ratio of these conversion weights is
approximately $50$.

On the discrete evaluation grid
$t\in\{0.1,0.3,0.5,0.7,0.9\}$ used in
\cref{tab:patch_sweep,tab:eps_threshold_app},
the corresponding averages are $17.53$ and
$0.33$, respectively.
The tabulated lower bounds additionally divide
by $D$ to match the reported per-dimension MSE.

\section{Additional experiments and details}
\label{sec:exp_details}

\subsection{Bottleneck lower bound: synthetic validation}
\label{sec:synthetic}
\definecolor{good}{HTML}{D4EDDA}
\definecolor{mid}{HTML}{FFF3CD}
\definecolor{bad}{HTML}{F8D7DA}
\definecolor{catastrophic}{HTML}{C0392B}

\begin{table}[t]
\caption{
Comparison of validation $x$-space MSE across prediction and loss choices on Imagenette.
For the U-Net without an information bottleneck, MSE varies primarily with the loss space.
For JiT-B with a bottleneck ($r=0.67$), MSE varies primarily with the prediction space.
The full results, including FID, are reported in \cref{tab:unet_results,tab:jit_results}.
}
\label{fig:loss_grid}
\centering
\small
\begin{tabular}{@{}c l ccc c ccc@{}}
\toprule
& & \multicolumn{3}{c}{\textbf{U-Net} (no bottleneck)} & & \multicolumn{3}{c}{\textbf{JiT-B} ($r{=}0.67$, bottleneck)} \\
\cmidrule(lr){3-5} \cmidrule(l){7-9}
& Pred$\backslash$Loss & $x$ & $\epsilon$ & $v$ & & $x$ & $\epsilon$ & $v$ \\
\midrule
& $x$
  & \cellcolor{mid}0.0481  & \cellcolor{good}0.0466  & \cellcolor{good}\textbf{0.0463}
  & & \cellcolor{good}0.050  & \cellcolor{good}\textbf{0.048}  & \cellcolor{good}0.048 \\
& $\epsilon$
  & \cellcolor{mid}0.0486  & \cellcolor{good}0.0471  & \cellcolor{good}0.0467
  & & \cellcolor{catastrophic}\color{white}\textbf{5.9}  & \cellcolor{catastrophic}\color{white}\textbf{5.8}  & \cellcolor{catastrophic}\color{white}\textbf{5.9} \\
& $v$
  & \cellcolor{mid}0.0476  & \cellcolor{good}0.0465  & \cellcolor{good}0.0463
  & & \cellcolor{bad}0.152  & \cellcolor{bad}0.152  & \cellcolor{bad}0.152 \\
\bottomrule
\end{tabular}
\end{table}

We validate the irreducible error lower bound (\cref{thm:nullspace_lb}) using the synthetic manifold setup with a controlled linear bottleneck.
Fix ambient dimension $D$ and intrinsic dimension $d \ll D$.
We sample a column-orthonormal matrix $P\in\R^{D\times d}$, draw $\hat x \sim p_{\hat x}$ from a mixture of Gaussians on $\R^d$, and embed as $x = P\hat x$.
With $\epsilon \sim \mathcal{N}(0,I_D)$ and $z_t = tx + (1{-}t)\epsilon$, the model accesses only $\rho = Ez_t \in \R^m$ through a trainable embedding $E\in\R^{m\times D}$.
We train all nine $(s,u)$ combinations and sweep $m$ from $1$ to $D$, decomposing the MSE into null-space ($\Pi_{\mathrm{null}}$) and range-space ($\Pi_{\mathrm{range}}$) components.
We report prediction MSE in each target's native space, together with its null-space and range-space components and convergence diagnostics.
As predicted in \cref{fig:bottleneck}, $\epsilon$- and $v$-prediction exhibit an irreducible error floor, while $x$-prediction achieves low error for all $m \geq d$.
For $\epsilon$, the null-space MSE matches $D{-}m$ precisely, confirming the theorem; for $v$, it lies on or above this floor, as \cref{cor:separation} requires.

\paragraph{Sweep over $D$, $d$, $m$, and the embedding $E$.}
We extend this setup along all three dimensions of \cref{thm:nullspace_lb,cor:separation}: $D \in \{16, 64, 128\}$, $d \in \{2, 8\}$, and $m$ from below $d$ up to $D$, for all three prediction types.
Each configuration is trained with $v$-loss for $2{,}000$ epochs, once with a learned $E$ and once with a random $E$ frozen at initialization, over two seeds, for $444$ runs in total.
We report squared error per vector with $t \sim \mathcal{U}(0,1)$, averaged over seeds; \cref{tab:toy_sweep} lists every configuration.
Three findings stand out.
First, the floor of \cref{thm:nullspace_lb} is exact in all three variables. Across the $62$ configurations with $m < D$, the null-space error of $\epsilon$-prediction lies between $0.997$ and $1.012$ times $D{-}m$.
$v$-prediction lies above the same floor in every configuration, as \cref{cor:separation} predicts. Its proof adds the posterior error of $\Pi_{\mathrm{null}} x$ to the $\epsilon$ floor, and with a learned $E$ and $m \geq d$ this excess stays between $12.8$ and $18.7$ for every $D$ and $m$.
Second, $x$-prediction depends on whether $m \geq d$ rather than on $D$ or $m$. With a learned $E$, its error stays between $11.4$ and $12.5$ whenever $m \geq d$ and rises to between $15.9$ and $17.1$ when $m < d$, so the transition sits at the injectivity condition (A4).
Third, a random $E$ separates the two mechanisms. The $\epsilon$ floor is the same for random and learned $E$ to within $1.5\%$, so it depends only on $\mathrm{rank}(E)$ and is information-theoretic rather than an artifact of training the embedding.
For $x$-prediction, a random $E$ is much worse near $m = d$, for example $24.9$ against $11.4$ at $D{=}128$ and $m{=}d{=}2$, and approaches the learned value as $m \to D$.
A learned $E$ thus finds the subspace-aligned solution already at the minimal rank $m = d$, consistent with the discussion after \cref{cor:separation} that end-to-end training pushes $E$ to preserve the data subspace.

\begin{table}[t!]
\caption{Synthetic sweep of \cref{thm:nullspace_lb,cor:separation} over ambient dimension $D$, intrinsic dimension $d$, and bottleneck rank $m$, with a learned or a frozen random embedding $E$ (BottleneckMLP, $v$-loss, $2$ seeds). The $\epsilon$ and $v$ columns give the null-space error of $\epsilon$- and $v$-prediction divided by the floor $D{-}m$; the $x$ columns give the total error of $x$-prediction. Errors are squared norms per vector with $t \sim \mathcal{U}(0,1)$. Rows with $m{=}D$ have no floor.}
\label{tab:toy_sweep}
\centering\footnotesize
\begin{tabular}{@{}rrr ccc cc@{}}
\toprule
& & & \multicolumn{2}{c}{$\epsilon$ null / $(D{-}m)$} & $v$ null / $(D{-}m)$ & \multicolumn{2}{c}{$x$ error} \\
\cmidrule(lr){4-5} \cmidrule(lr){6-6} \cmidrule(l){7-8}
$D$ & $d$ & $m$ & learned & random & learned & learned & random \\
\midrule
16 & 2 & 1 & 1.001 & 1.001 & 1.96 & 17.0 & 23.5 \\
16 & 2 & 2 & 1.002 & 1.003 & 2.04 & 11.5 & 22.7 \\
16 & 2 & 4 & 1.003 & 1.003 & 2.19 & 11.4 & 16.0 \\
16 & 2 & 8 & 1.006 & 1.005 & 2.80 & 11.6 & 13.0 \\
16 & 2 & 12 & 0.997 & 1.012 & 4.64 & 11.6 & 12.2 \\
16 & 2 & 16 & --- & --- & --- & 11.7 & 11.6 \\
16 & 8 & 4 & 1.000 & 1.003 & 2.22 & 16.0 & 18.2 \\
16 & 8 & 8 & 1.007 & 1.001 & 2.80 & 12.2 & 16.1 \\
16 & 8 & 12 & 1.006 & 1.000 & 4.21 & 11.6 & 12.8 \\
16 & 8 & 16 & --- & --- & --- & 11.5 & 11.5 \\
\midrule
64 & 2 & 1 & 1.002 & 1.003 & 1.23 & 17.1 & 21.9 \\
64 & 2 & 2 & 1.003 & 1.003 & 1.24 & 11.7 & 18.8 \\
64 & 2 & 4 & 1.002 & 1.003 & 1.25 & 11.7 & 16.7 \\
64 & 2 & 16 & 1.004 & 1.004 & 1.30 & 11.6 & 14.8 \\
64 & 2 & 32 & 1.006 & 1.005 & 1.46 & 11.8 & 13.0 \\
64 & 2 & 48 & 1.007 & 1.006 & 1.90 & 11.8 & 12.2 \\
64 & 2 & 64 & --- & --- & --- & 11.9 & 11.3 \\
64 & 8 & 4 & 1.003 & 1.003 & 1.25 & 15.9 & 18.8 \\
64 & 8 & 8 & 1.004 & 1.003 & 1.27 & 12.2 & 17.5 \\
64 & 8 & 16 & 1.004 & 1.005 & 1.37 & 11.7 & 15.6 \\
64 & 8 & 32 & 1.007 & 1.007 & 1.58 & 11.6 & 13.5 \\
64 & 8 & 48 & 1.007 & 1.006 & 2.13 & 11.6 & 12.3 \\
64 & 8 & 64 & --- & --- & --- & 11.8 & 11.4 \\
\midrule
128 & 2 & 1 & 1.004 & 1.004 & 1.12 & 16.8 & 24.6 \\
128 & 2 & 2 & 1.004 & 1.004 & 1.12 & 11.4 & 24.9 \\
128 & 2 & 4 & 1.004 & 1.004 & 1.12 & 11.6 & 22.9 \\
128 & 2 & 32 & 1.006 & 1.006 & 1.15 & 11.6 & 14.0 \\
128 & 2 & 64 & 1.007 & 1.008 & 1.23 & 11.7 & 12.7 \\
128 & 2 & 96 & 1.012 & 1.010 & 1.43 & 11.6 & 12.0 \\
128 & 2 & 128 & --- & --- & --- & 11.5 & 11.6 \\
128 & 8 & 4 & 1.004 & 1.004 & 1.13 & 15.9 & 19.6 \\
128 & 8 & 8 & 1.005 & 1.004 & 1.13 & 12.5 & 18.1 \\
128 & 8 & 16 & 1.005 & 1.005 & 1.17 & 11.7 & 17.2 \\
128 & 8 & 32 & 1.005 & 1.006 & 1.19 & 11.7 & 15.0 \\
128 & 8 & 64 & 1.006 & 1.008 & 1.28 & 11.6 & 13.2 \\
128 & 8 & 96 & 1.009 & 1.011 & 1.56 & 11.7 & 12.4 \\
128 & 8 & 128 & --- & --- & --- & 11.6 & 11.7 \\
\bottomrule
\end{tabular}
\end{table}

\subsection{Bottleneck threshold sweep}
\label{sec:bottle_sweep}

\paragraph{BottleneckViT and BottleneckMLP.}
BottleneckViT is a simplified Vision Transformer
used as a computationally cheaper proxy for JiT-B
in bottleneck-width sweeps.

Its patch embedding follows JiT's
\texttt{BottleneckPatchEmbed}, but inserts a GELU
between the two convolutions.
Importantly, the first convolution is a linear
mapping from the noisy patch to a
$d_{\mathrm{bn}}$-dimensional representation.
All subsequent computations, including the GELU,
depend on the input only through this representation.
Thus, under compression ($d_{\mathrm{bn}}<
d_{\mathrm{patch}}$), the architecture satisfies
assumption (A3) of \cref{thm:nullspace_lb},
provided the first mapping has full row rank.

BottleneckViT uses fewer transformer blocks and
a smaller hidden width than JiT-B, giving
approximately 22M parameters versus JiT-B's 130M.
This enables systematic bottleneck-width sweeps
at substantially lower computational cost.

BottleneckMLP is the analogous synthetic-data
variant, using an MLP backbone with a linear
input bottleneck.

\Cref{tab:bn_sweep_app,tab:eps_threshold_app} show the full bottleneck-ratio sweep on BottleneckViT.

\begin{table}[h]
\caption{Bottleneck dimension sweep on CIFAR-10 (BottleneckViT, $d_\mathrm{patch} = 48$).}
\label{tab:bn_sweep_app}
\centering\small
\begin{tabular}{@{}r r c c r@{}}
\toprule
$d_\mathrm{bn}$ & Ratio $r$ & $x$-pred MSE & $v$-pred MSE & Gap \\
\midrule
32  & 0.67 & 0.054 & 0.158 & $+191\%$ \\
64  & 1.33 & 0.054 & 0.054 & $+0.2\%$ \\
128 & 2.67 & 0.054 & 0.054 & $+0.0\%$ \\
256 & 5.33 & 0.054 & 0.054 & $+0.0\%$ \\
\bottomrule
\end{tabular}
\end{table}

\begin{table}[h]
\caption{
$\epsilon$-prediction recovery with increasing bottleneck
width on Imagenette ($d_\mathrm{patch}=192$).
Results use BottleneckViT with $\epsilon$-loss,
500 training epochs, and seed 42.
Validation $x$-space MSE is averaged over
$t\in\{0.1,0.3,0.5,0.7,0.9\}$.
The floor is $17.53(D-m)_+/D$, obtained from
\cref{cor:pred_threshold} on this evaluation grid.
}
\label{tab:eps_threshold_app}
\centering\small
\begin{tabular}{@{}r r c c r c@{}}
\toprule
$d_\mathrm{bn}$ & Ratio $r$ & $x$-pred MSE &
$\epsilon$-pred MSE & Gap & Floor \\
\midrule
32  & 0.17 & 0.049 & 14.61 & $297\times$ & 14.61 \\
256 & 1.33 & 0.048 & 0.095 & $+98\%$ & --- \\
384 & 2.00 & 0.048 & 0.056 & $+17\%$ & --- \\
512 & 2.67 & 0.048 & 0.053 & $+9\%$ & --- \\
768 & 4.00 & 0.048 & 0.050 & $+2.6\%$ & --- \\
\bottomrule
\end{tabular}
\end{table}

\paragraph{Patch size at fixed embedding width.}
\Cref{thm:nullspace_lb} predicts that the floor depends on the patch through $D{-}m = p^2 C - d_{\mathrm{bn}}$.
Increasing the patch size $p$ at fixed $d_{\mathrm{bn}}$ should therefore degrade $\epsilon$- and $v$-prediction monotonically, while leaving $x$-prediction unaffected as long as $d_{\mathrm{bn}}$ exceeds the intrinsic dimension of a patch.
We test this with BottleneckViT on CIFAR-10 at $p \in \{2, 4, 8\}$ and fixed $d_{\mathrm{bn}} = 32$, so that $D = 3p^2$ ranges from $12$ to $192$ under the same dataset and training recipe: $v$-loss, $200$ epochs, seed 42.
\Cref{tab:patch_sweep} reports validation MSE averaged over $t \in \{0.1, 0.3, 0.5, 0.7, 0.9\}$.
At $p{=}2$ the embedding is wider than the patch, so there is no bottleneck and the three prediction types match.
As $p$ grows, $x$-prediction stays flat at $0.054$ to $0.056$, while $\epsilon$- and $v$-prediction degrade monotonically.
The $\epsilon$-space error of $\epsilon$-prediction lands on the floor at both bottlenecked sizes. Averaged over $t$ it is $0.353$ against $0.333$ and $0.842$ against $0.833$, and for $t \leq 0.5$ it matches the floor to within $1\%$.
Through \cref{prop:conversion_amp}, the floor then predicts the $x$-space error of $\epsilon$-prediction, as in \cref{cor:pred_threshold}. The mean amplification $(1{-}t)^2/t^2$ over the evaluation grid is $17.53$, so the prediction $17.53\,(D{-}m)/D$ is $5.84$ at $p{=}4$ and $14.61$ at $p{=}8$, against measured values of $5.85$ and $14.62$.
$v$-prediction lies above its corresponding bound $0.33\,(D{-}m)/D$, which is $0.11$ and $0.28$, because its floor also contains the $x$ component.
The same values recur across setups. At $r{=}0.67$ the $\epsilon$-prediction error of $5.85$ equals that of JiT-B/8 on Imagenette (\cref{tab:jit_results}), and at $r{=}0.17$ the value $14.62$ matches JiT-B/16 on ImageNet-256 (\cref{tab:imagenet256}).
Since the error sits on the floor and the amplification depends only on the evaluation grid, the bottleneck ratio governs this error, not the resolution or the architecture.

\begin{table}[h]
\caption{Patch-size sweep on CIFAR-10 (BottleneckViT, $d_{\mathrm{bn}}{=}32$, $v$-loss, $200$ epochs, seed 42). Validation MSE per dimension, averaged over $t \in \{0.1, 0.3, 0.5, 0.7, 0.9\}$. The first three error columns are in $x$-space; the last is the $\epsilon$-space error of $\epsilon$-prediction, to be compared with the floor $(D{-}m)/D$.}
\label{tab:patch_sweep}
\centering\small
\begin{tabular}{@{}rrcc ccc c@{}}
\toprule
& & & & \multicolumn{3}{c}{$x$-space error} & $\epsilon$-space error \\
\cmidrule(lr){5-7} \cmidrule(l){8-8}
$p$ & $D{=}3p^2$ & $r$ & $(D{-}m)/D$ & $x$-pred & $\epsilon$-pred & $v$-pred & $\epsilon$-pred \\
\midrule
2 & 12 & 2.67 & --- & 0.054 & 0.055 & 0.054 & 0.060 \\
4 & 48 & 0.67 & 0.333 & 0.054 & 5.85 & 0.158 & 0.353 \\
8 & 192 & 0.17 & 0.833 & 0.056 & 14.62 & 0.321 & 0.842 \\
\bottomrule
\end{tabular}
\end{table}

\subsection{Compression vs.\ post-bottleneck nonlinearity}
\label{sec:non-linear}

We examine whether introducing nonlinear processing after the initial linear bottleneck affects the prediction-space asymmetry.
Specifically, we insert a GeLU between JiT's two patch-embedding convolutions.
In both variants, the first convolution compresses each noisy patch into a $d_{\mathrm{bn}}$-dimensional representation, and all subsequent computations depend only on this representation.
Thus, the linear bottleneck assumption of \cref{thm:nullspace_lb} remains satisfied under compression.

As shown in \cref{tab:nonlinear_bottleneck}, $\epsilon$-prediction exhibits similarly large errors under compression in both variants, while its error decreases substantially when compression is removed.
These results demonstrate that the prediction-space asymmetry persists despite the additional nonlinearity after the initial linear compression.

\begin{table}[h]
\centering
\small
\caption{Effect of post-bottleneck nonlinearity on Imagenette.
A $2{\times}2$ factorial design compares JiT's original linear patch embedding with a variant inserting a GeLU between its two convolutions, under compressed and uncompressed configurations.
Cells show test MSE; numbers in parentheses give the $\epsilon$-prediction gap relative to $x$-prediction.}
\label{tab:nonlinear_bottleneck}
\begin{tabular}{@{}l cc cc@{}}
\toprule
& \multicolumn{2}{c}{$d_{\mathrm{bn}}{=}128$ (compression)} & \multicolumn{2}{c}{$d_{\mathrm{bn}}{=}256$ (no compr.)} \\
\cmidrule(lr){2-3} \cmidrule(l){4-5}
& $x$-pred & $\epsilon$-pred & $x$-pred & $\epsilon$-pred \\
\midrule
Linear & $0.048$ & $5.85$ ($120{\times}$) & $0.048$ & $0.051$ ($6\%$) \\
GeLU   & $0.049$ & $5.9\phantom{0}$ ($120{\times}$) & $0.049$ & $0.059$ ($21\%$) \\
\bottomrule
\end{tabular}
\end{table}

\subsection{Generated sample comparison}

\begin{figure}[h]
    \centering
    \includegraphics[width=\linewidth]{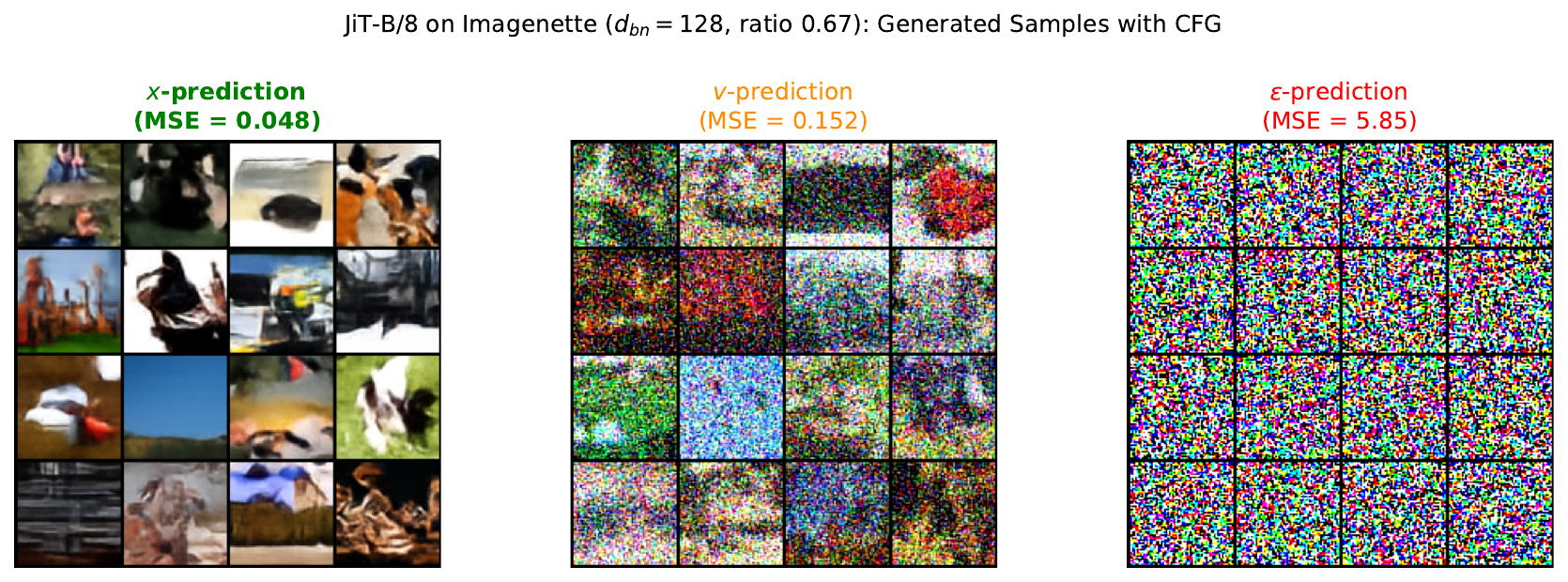}
    \caption{Generated samples from JiT-B/8 ($d_{\mathrm{bn}}{=}128$, ratio $0.67$) on Imagenette with classifier-free guidance. $x$-prediction produces recognizable images; $v$-prediction is heavily degraded; $\epsilon$-prediction produces pure noise. All three use the same 130M-parameter architecture and $100$ Heun steps; $x$- and $\epsilon$-prediction use $\epsilon$-loss and $v$-prediction uses $v$-loss.}
    \label{fig:samples}
\end{figure}

To complement the quantitative MSE and FID gaps reported in \cref{tab:jit_results}, we visualize the qualitative effect of prediction-space choice under a fixed bottleneck.
\Cref{fig:samples} shows samples generated by JiT-B/8 with $d_{\mathrm{bn}}{=}128$ at ratio $0.67$ on Imagenette using classifier-free guidance, with $100$ Heun steps. All three rows use the same architecture and training recipe. $x$- and $\epsilon$-prediction are trained with $\epsilon$-loss and $v$-prediction with $v$-loss; under this bottleneck the loss choice changes MSE by at most $3.3\%$ within a prediction type (\cref{tab:jit_results}), so the rows differ mainly in the prediction target.
The visual ordering tracks the MSE ordering precisely: $x$-prediction yields recognizable class-conditional images, $v$-prediction yields heavily degraded but still structured samples, and $\epsilon$-prediction collapses to pure noise.
This corroborates \cref{thm:nullspace_lb}: the irreducible null-space error of $D-m$ for $\epsilon$ propagates through the ODE sampler to produce samples with no learnable structure, while $x$-prediction, free of the null-space penalty under the manifold assumption, recovers usable samples even at the same bottleneck width.

\subsection{JiT-B/4 convergence behavior}

\begin{figure}[h]
    \centering
    \includegraphics[width=\linewidth]{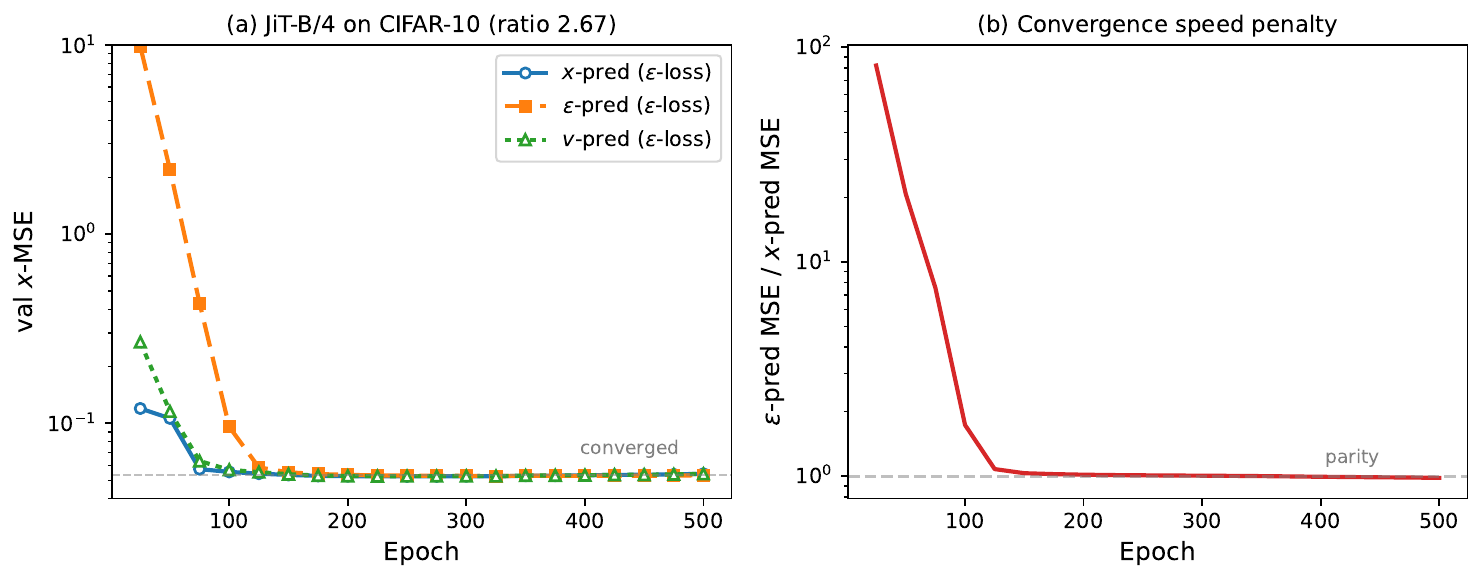}
    \caption{Loss curves on JiT-B/4 CIFAR-10 ($r{=}2.67$, $\epsilon$-loss). \textbf{(a)}~$\epsilon$-prediction starts roughly 80-fold worse than $x$-prediction but reaches parity by epoch 200. \textbf{(b)}~The MSE ratio decays from 80-fold to roughly unity, suggesting that the bottleneck effect at $r \geq 1$ is a convergence penalty rather than an irreducible error.}
    \label{fig:convergence_speed}
\end{figure}

\Cref{fig:convergence_speed} reports per-epoch validation MSE for JiT-B/4 on CIFAR-10 at bottleneck ratio $r{=}2.67$, trained for 500 epochs with $\epsilon$-loss, alongside an $x$-prediction baseline trained under the same setup.
Panel (a) plots the absolute loss curves; panel (b) shows their ratio.
This single training run is the evidence for the convergence-penalty framing in \cref{sec:loss_analysis}: it shows that, on this configuration, the prediction-type gap closes with continued training rather than persisting as an information-theoretic floor.
We have not measured this systematically across $r$, datasets, or architectures.
The ImageNet-256 study (\cref{sec:imagenet256}) adds one more point at the regime boundary: at $r{=}1.0$ and $600$ epochs, the $\epsilon$-prediction denoising gap narrows to a factor of two without fully closing, and the FID gap from conversion amplification persists regardless.

\subsection{Numerical evaluation of the loss-ordering integral}
\label{sec:loss_integral}

We evaluate the integral $J_\kappa(w_{\mathrm{eff}})$ of \cref{thm:loss_bound} for the three effective weights under the logit-normal time distributions used in training.
A logit-normal density vanishes at both endpoints, so $J_\kappa$ diverges on $(0,1)$. We therefore integrate over $t \in [0.05, 0.95]$. This cutoff is a modeling choice: it equals the conversion clamp $t_\epsilon = 0.05$ of the JiT recipe, but the training time distribution itself covers $(0,1)$.
\Cref{tab:loss_integral} normalizes each integral by its $v$-loss value.
With constant $\kappa$, which ignores assumption (iv), the $x$-loss integral is $126$ to $256$ times the $v$-loss integral, while the $\epsilon$-loss integral is $1.5$ to $11.4$ times.
The excess of $\epsilon$-loss comes from low $t$, where its weight $t^2$ vanishes.
Restricting the sensitivity to high $t$, $\kappa(t) = \mathbf{1}[t > 1/2]$, brings the $\epsilon$-loss integral within a factor of $1.43$ of the $v$-loss integral and leaves the $x$-loss integral more than two orders of magnitude larger, which is the ordering~\eqref{eq:loss_ordering}.
For integration cutoffs between $0.02$ and $0.1$, the $x$-loss integral remains the largest, by a factor of at least $28$.

\begin{table}[h]
\caption{The loss-ordering integral $J_\kappa(w_{\mathrm{eff}})$ of \cref{thm:loss_bound} over $t \in [0.05, 0.95]$, normalized by its $v$-loss value. Effective weights are $1$, $t^2$, and $(1{-}t)^2$ for $v$-, $\epsilon$-, and $x$-loss.}
\label{tab:loss_integral}
\centering\small
\begin{tabular}{@{}l l ccc@{}}
\toprule
Time distribution & Sensitivity $\kappa(t)$ & $v$-loss & $\epsilon$-loss & $x$-loss \\
\midrule
\multirow{2}{*}{logit-normal$(-0.5, 1.0)$, U-Net} & $1$ & $1$ & $11.4$ & $125.9$ \\
 & $\mathbf{1}[t > 1/2]$ & $1$ & $1.43$ & $148.5$ \\
\midrule
\multirow{2}{*}{logit-normal$(-0.8, 0.8)$, JiT} & $1$ & $1$ & $1.49$ & $255.9$ \\
 & $\mathbf{1}[t > 1/2]$ & $1$ & $1.17$ & $257.0$ \\
\bottomrule
\end{tabular}
\end{table}

\subsection{Loss ordering on simple distributions}

The ordering~\eqref{eq:loss_ordering} relies on assumption (iv) of \cref{thm:loss_bound}, that generation quality depends disproportionately on high-$t$ accuracy.
This holds for natural images, where fine-grained details such as textures and edges are resolved near $t{=}1$ and dominate FID.
However, for simple distributions such as 2D synthetic datasets, the denoising task is approximately equally difficult at all timesteps.
In this regime, our synthetic experiments show that the loss ordering partially \emph{reverses}: diagonal losses ($u{=}s$, uniform weight) become near-optimal, and $\epsilon$-loss can underperform $x$-loss.
This is consistent with the analysis: when denoising difficulty is uniform across $t$, concentrating capacity on any particular timestep range cannot help.

\subsection{ImageNet-256 scale validation}
\label{sec:imagenet256}

To confirm that both regimes persist at the resolution and dataset scale of modern generative models, we train the full JiT-B/16 architecture on class-conditional ImageNet-1k at $256{\times}256$ ($1.28$M images, $1000$ classes), where each patch has $d_{\mathrm{patch}} = 16^2 {\cdot} 3 = 768$.
We run the complete $3{\times}3$ design at $d_{\mathrm{bn}}{=}128$ (ratio $r{=}0.17$, strong compression) for $100$ epochs each, and a three-configuration control at $d_{\mathrm{bn}}{=}768$ (ratio $r{=}1.0$, no compression) trained to $600$ epochs.
All runs use batch size $1024$, AdamW ($\beta_1{=}0.9$, $\beta_2{=}0.95$, lr $10^{-4}$, constant after a $5$-epoch warmup), logit-normal time sampling ($\mu{=}{-}0.8$, $\sigma{=}0.8$), label dropout $0.1$, velocity clamp $t_{\epsilon}{=}0.05$, and EMA decay $0.9999$, matching the JiT recipe of the Imagenette experiments.
FID is computed with clean-fid~\citep{parmar2022cleanfid} over $50$K samples (Heun, $50$ steps, CFG $2.9$) against the full training set.
\Cref{tab:imagenet256} reports the results.

Both phenomena from the smaller-scale experiments reproduce. Under compression ($r{=}0.17$), the prediction-space dichotomy is stark: $x$-prediction attains val $x$-MSE $0.023$ and FID $20$--$37$ across all loss types, $v$-prediction degrades to MSE $0.29$ and FID ${\approx}200$, and $\epsilon$-prediction collapses to MSE $14.6$ and FID ${\approx}420$.
The $\epsilon$-prediction error in $\epsilon$-space sits at $0.834$--$0.837$ across the three loss types, within $0.5\%$ of the closed-form floor $(D{-}m)/D = 640/768 = 0.833$ of \cref{thm:nullspace_lb}, a direct quantitative confirmation of the bottleneck lower bound at scale.

Removing the bottleneck ($r{=}1.0$) sharpens the central message of the paper. We train all three control configurations to the full $600$-epoch budget so that the comparison is at convergence rather than confounded by training time. The val $x$-MSE spread collapses from $600$-fold under compression to within a factor of two ($0.022$--$0.043$), confirming that information is no longer lost; yet the FID gap remains enormous ($5.7$ for $x$-pred, $140.6$ for $v$-pred, $330.6$ for $\epsilon$-pred).
This is the dual dichotomy: once the architecture can represent the target, denoising quality (MSE) nearly equalizes, while prediction type still dominates generation quality (FID).
For $\epsilon$-prediction, the per-timestep analysis below locates the gap at low $t$, where the $\epsilon \to v$ conversion of \cref{prop:conversion_amp} is ill-conditioned.
$v$-prediction needs no conversion for ODE sampling, so \cref{prop:conversion_amp} does not explain its gap. Its velocity error stays within $30\%$ of the $x$-model's at every $t$ we measure and is lower at $t{=}0.95$ (\cref{tab:common_space}), and we leave the cause of its FID gap open.
Crucially, the gap is \emph{not} an undertraining artifact: from $100$ to $600$ epochs the $x$-prediction FID falls sharply ($24.8 \to 5.7$, reaching the range of published ImageNet-$256$ models~\citep{li2025denoise}), while $\epsilon$-prediction barely moves ($340.2 \to 330.6$) and $v$-prediction improves only modestly ($173.5 \to 140.6$). At convergence, $v$-prediction remains ${\approx}25{\times}$ worse than $x$-prediction in FID despite a near-identical $x$-MSE ($0.026$ vs.\ $0.022$), so the FID-vs-MSE decoupling is not a transient of finite training or an artifact of the bottleneck.

\Cref{fig:pertimestep} makes the mechanism explicit. Measuring per-timestep error on the converged $r{=}1.0$ checkpoints, $\epsilon$-prediction's \emph{velocity}-space error, the quantity the ODE integrates, grows sharply as $t \to 0$, reaching $0.44$ at $t{=}0.06$ and $1.07$ at $t{=}0.02$ versus ${\approx}0.065$ at mid-range. Above $t_\epsilon$ this is the low-$t$ blow-up predicted by \cref{prop:conversion_amp}, and below $t_\epsilon$ the conversion clamp described next also contributes; $x$- and $v$-prediction stay bounded.
Following the JiT recipe, every conversion clamps its denominator $t$ or $1{-}t$ at $t_\epsilon = 0.05$.
The sampler starts at $t{=}0$, so the first steps of $\epsilon$-prediction, at $t = 0$, $0.02$, and $0.04$, fall in the clamped region, as do the last steps of $x$-prediction at $t = 0.96$ and $0.98$.
There the clamp caps the amplification at $1/t_\epsilon$, but even an exact network output then yields a velocity shrunk by the factor $t/t_\epsilon$ for $\epsilon$-prediction, or $(1{-}t)/t_\epsilon$ for $x$-prediction.
The $\epsilon$-prediction FID gap therefore reflects the $\epsilon \to v$ conversion together with the clamp that regularizes it, and we do not separate the two contributions.

\paragraph{Comparing separately trained models.}
Different models may have different prediction errors before
conversion.
We therefore compare their predictions in velocity space,
using the same target $v=x-\epsilon$ for all three models.
This is also the quantity used by the ODE sampler.

\begin{remark}[Comparing prediction errors in velocity space]
\label{rem:excess}
For a model predicting $u\in\{x,\epsilon,v\}$, let
$u^\star=\E[u\mid z_t,t]$ be the Bayes predictor and
$e_u=\hat u-u^\star$ the error relative to it.
The class label is implicit in the conditioning of the
Bayes predictors.
These predictors map exactly onto one another under the
affine conversions in \cref{thm:weighted_reduction}.

Let $\hat v_u$ denote the velocity estimate obtained from
$\hat u$.
At each fixed $t$ where denominator clamping is inactive,
\[
\E\!\left[\|\hat v_u-v\|^2\mid t\right]
=
\underbrace{
  \E\!\left[\|v^\star-v\|^2\mid t\right]
}_{\text{common Bayes risk}}
+
w_{v\mid u}(t)\,
\E\!\left[\|e_u\|^2\mid t\right],
\]
where $w_{v\mid u}(t)$ is given in \cref{tab:weights}.
The cross term vanishes because
$\E[v-v^\star\mid z_t,t]=0$ and
$\hat v_u-v^\star$ is determined by the model's inputs.

All three prediction types share the same Bayes risk in
velocity space.
Differences in their expected velocity MSEs therefore equal
differences in their excess MSEs.
The converted excess MSE depends on both the conversion
weight and the excess MSE before conversion.
The same decomposition holds for MSE per dimension after
normalizing all terms by the number of output coordinates.
\end{remark}

\Cref{tab:common_space} reports empirical estimates of these
velocity-space MSEs.
At $t=0.06$, the $\epsilon$-model's MSE is $0.0016$ before
conversion and $0.443$ in velocity space, where the conversion
weight is $w_{v\mid\epsilon}(t)=1/t^2\approx278$.
The $x$-model's velocity-space MSE is $0.106$.
The observed difference of $0.337$ estimates the corresponding
difference in excess MSE.

The ordering between these two models reverses at $t=0.95$:
the $\epsilon$-model has a lower velocity-space MSE than
the $x$-model ($0.337$ versus $0.462$).
Thus, the comparison depends on the learned prediction errors
as well as the conversion weights.
The table excludes timesteps where denominator clamping
changes the conversion formulas.

\begin{table}[h]
\caption{Per-timestep error of the three separately trained JiT-B/16 models on ImageNet-256 ($r{=}1.0$, $600$ epochs, $v$-loss) in the common velocity space, together with the $\epsilon$-model's native error $\|\delta_\epsilon\|^2 = t^2\|\delta v\|^2$. Mean squared error per dimension against sampled targets, which includes the Bayes risk, over $2{,}560$ validation images.}
\label{tab:common_space}
\centering\small
\begin{tabular}{@{}c ccc c@{}}
\toprule
& \multicolumn{3}{c}{Velocity-space error} & Native error \\
\cmidrule(lr){2-4} \cmidrule(l){5-5}
$t$ & $x$-model & $\epsilon$-model & $v$-model & $\epsilon$-model \\
\midrule
0.06 & 0.106 & 0.443 & 0.116 & 0.0016 \\
0.10 & 0.081 & 0.198 & 0.093 & 0.0020 \\
0.50 & 0.051 & 0.067 & 0.066 & 0.017 \\
0.90 & 0.189 & 0.197 & 0.197 & 0.160 \\
0.95 & 0.462 & 0.337 & 0.333 & 0.304 \\
\bottomrule
\end{tabular}
\end{table}

\paragraph{Stochastic sampling.}
\Cref{prop:conversion_amp} concerns the drift field, not the integrator.
Consider the reverse samplers that share the marginals of the probability-flow ODE: for any schedule $w(t) \geq 0$,
\begin{equation}
\label{eq:sde_family}
dz = \big[\hat v(z,t) + w(t)\,\hat s(z,t)\big]\,dt + \sqrt{2w(t)}\,dW_t,
\end{equation}
where $\hat s$ estimates the score $\nabla \log p_t$ and $w \equiv 0$ recovers the ODE.
Both drift terms are affine conversions of the same network output.
For an $\epsilon$-prediction model, $\hat v = (z_t - \hat\epsilon)/t$ and $\hat s = -\hat\epsilon/(1{-}t)$, so a native error $\delta_\epsilon$ gives
\[
\delta v = -\frac{\delta_\epsilon}{t}, \qquad \delta s = -\frac{\delta_\epsilon}{1-t}.
\]
The low-$t$ amplification therefore stays in the drift for every choice of $w$.
Injected noise can change how a drift error affects the sample distribution, but it does not remove a systematic error in the field.
We test this on the $600$-epoch $r{=}1.0$ checkpoints with Euler--Maruyama sampling of~\eqref{eq:sde_family}, $w(t) = 1{-}t$, and $50$ steps, against the Heun ODE with the same step count. Heun evaluates the network twice per step, so the ODE uses $99$ network evaluations against $50$ for the SDE.
Both samplers use a matched $4$K-sample FID protocol, which is upward-biased relative to the $50$K numbers in \cref{tab:imagenet256} but comparable across samplers.
\Cref{tab:sde} shows that $\epsilon$-prediction does not recover under stochastic sampling, while $x$-prediction changes only modestly.
Scaling the schedule to $w(t) = 4(1{-}t)$ and $16(1{-}t)$ keeps the $\epsilon$-prediction FID between $328$ and $372$.
The FID gap thus originates in a drift-field error that the ODE and this SDE share.
This conclusion is limited to samplers that use the same conversion. Samplers that avoid it, such as DDPM ancestral sampling in native $\epsilon$-space, also avoid the amplification (\cref{sec:loss_analysis}).

\begin{table}[h]
\caption{ODE versus SDE sampling on the $600$-epoch ImageNet-256 checkpoints ($r{=}1.0$, $v$-loss). FID over $4$K samples, $50$ steps, CFG $2.9$. The SDE is Euler--Maruyama on~\eqref{eq:sde_family} with $w(t) = 1{-}t$.}
\label{tab:sde}
\centering\small
\begin{tabular}{@{}l cc@{}}
\toprule
FID ($4$K samples) & ODE (Heun) & SDE \\
\midrule
$x$-prediction & 14.5 & 17.1 \\
$\epsilon$-prediction & 333.4 & 330.2 \\
\bottomrule
\end{tabular}
\end{table}

\section{Relation to EDM preconditioning}
\label{sec:edm}

EDM~\citep{karras2022edm} writes the denoiser as
\[
D_\theta(\tilde z; \sigma) = c_{\mathrm{skip}}(\sigma)\,\tilde z + c_{\mathrm{out}}(\sigma)\,F_\theta\big(c_{\mathrm{in}}(\sigma)\,\tilde z;\, \sigma\big),
\]
\[
c_{\mathrm{skip}} = \frac{\sigma_{\mathrm{data}}^2}{\sigma^2 + \sigma_{\mathrm{data}}^2},
\qquad
c_{\mathrm{out}} = \frac{\sigma\,\sigma_{\mathrm{data}}}{\sqrt{\sigma^2 + \sigma_{\mathrm{data}}^2}},
\]
where $\tilde z = x + \sigma n$ with $n \sim \mathcal{N}(0, \Id_D)$.
Our interpolant takes this form with $\tilde z = z_t/t$ and $\sigma = (1{-}t)/t$.

EDM lies outside our static $3{\times}3$ grid for two reasons.
First, its network target changes continuously with the noise level. As $\sigma \to \infty$, $c_{\mathrm{skip}} \to 0$ and $F_\theta$ targets the rescaled data $x/\sigma_{\mathrm{data}}$. As $\sigma \to 0$, $c_{\mathrm{skip}} \to 1$ and $F_\theta$ targets a residual close to $-n$, an $\epsilon$-like quantity.
Second, the denoiser includes the full-rank skip path $c_{\mathrm{skip}}(\sigma)\,\tilde z$.

Our framework still applies to each ingredient.
An error $\delta_F$ in the network output enters the $x$-space estimate as $c_{\mathrm{out}}(\sigma)\,\delta_F$, so $c_{\mathrm{out}}^2$ plays the role of the amplification factor in \cref{prop:conversion_amp}.
In the same coordinates, that proposition gives the factor $\sigma^2$ for $\epsilon$-prediction, which is unbounded at high noise, and $1$ for $x$-prediction, which does not vanish at low noise.
EDM's factor satisfies $c_{\mathrm{out}}^2 \leq \min(\sigma^2, \sigma_{\mathrm{data}}^2)$ and is bounded at both ends.
EDM's loss weight $\lambda(\sigma) = 1/c_{\mathrm{out}}(\sigma)^2$ makes the effective weight on $\delta_F$ uniform over noise levels, which is a loss-space reweighting in the sense of \cref{thm:weighted_reduction}.
On the $x$-space error, $\lambda(\sigma) = \sigma^{-2} + \sigma_{\mathrm{data}}^{-2}$, our $\epsilon$-loss weight $w_{\epsilon\mid x} = t^2/(1{-}t)^2 = \sigma^{-2}$ plus a constant.
The skip term bypasses any compressed representation, so the final denoiser is not a function of $\rho = E z_t$ alone. Assumption (A3) then fails, and the $D{-}m$ floor of \cref{thm:nullspace_lb} does not apply.

We therefore expect EDM preconditioning to mitigate the bottleneck failure of $\epsilon$- and $v$-prediction, but we have neither a formal result nor a controlled experiment that establishes this.
A direct $c_{\mathrm{skip}}$ ablation on a bottlenecked ViT is an important next step.

\end{document}